\documentclass[preprint,12pt]{elsarticle}

\usepackage{caption}
\usepackage{epsfig}
\usepackage{graphicx}
\usepackage{amsmath, amssymb}
\usepackage{makecell}
\usepackage{multirow}
\usepackage{float}
\usepackage{subfigure}
\usepackage{booktabs}
\usepackage[misc]{ifsym}
\usepackage[dvipsnames]{xcolor}

\usepackage{amssymb}
\usepackage{bibleref}
\usepackage{natbib}
\usepackage{color}

\usepackage[colorlinks=True,
            linkcolor=red,
            filecolor=red,
            anchorcolor=red,
            citecolor=blue]{hyperref}

\journal{Neurocomputing}

\begin{document}

\begin{frontmatter}

\title{UniInst: Unique Representation for End-to-End Instance Segmentation}


\author[label1]{Yimin Ou\corref{EqualContribution}}
\author[label1]{Rui Yang\corref{EqualContribution}}
\author[label1]{Lufan Ma}
\author[label2]{Yong Liu\corref{CorrespondingAuthor}}
\author[label1]{Jiangpeng Yan}
\author[label2]{Shang Xu}
\author[label2]{Chengjie Wang}
\author[label1]{Xiu Li\corref{CorrespondingAuthor}}

\address[label1]{Tsinghua Shenzhen International Graduate School, Tsinghua University, Shenzhen, Guangdong, 518055, China}
\address[label2]{Tencent, Shenzhen, Guangdong, 518054, China}
            

\cortext[EqualContribution]{Equal Contribution}
\cortext[CorrespondingAuthor]{Corresponding Author}


\begin{abstract}
Existing instance segmentation methods have achieved impressive performance but still suffer from a common dilemma: redundant representations (e.g., multiple boxes, grids, and anchor points) are inferred for one instance, which leads to multiple duplicated predictions. Thus, mainstream methods usually rely on a hand-designed non-maximum suppression (NMS) post-processing step to select the optimal prediction result, consequently hindering end-to-end training.
To address this issue, we propose a box-free and NMS-free end-to-end instance segmentation framework, termed \textbf{UniInst}, which yields only one unique representation for each instance. Specifically, we design an instance-aware one-to-one assignment scheme, namely \textbf{O}nly \textbf{Y}ield \textbf{O}ne \textbf{R}epresentation (\textbf{OYOR}). It dynamically assigns one unique representation to each instance according to the matching quality between predictions and ground truths.
Then, a novel prediction re-ranking strategy is elegantly integrated into the framework to address the misalignment between the classification score and mask quality, enabling the learned representation to be more discriminative.
With these techniques, our UniInst, the first FCN-based box-free and NMS-free end-to-end instance segmentation framework, achieves competitive performance, e.g., \textbf{39.0 mask AP} using ResNet-50-FPN and \textbf{40.2 mask AP} using ResNet-101-FPN, against mainstream methods on COCO \textit{test-dev2017}.
Moreover, the proposed instance-aware method is robust to occlusion scenes because of non-dependent on the box and NMS. It therefore outperforms common baselines by remarkable mask AP on the heavily-occluded OCHuman benchmark.
Code is available at \url{https://github.com/b03505036/UniInst}.
\end{abstract}

\begin{keyword}
Instance Segmentation; End-to-End Instance Segmentation; Fully Convolutional Networks
\end{keyword}
\end{frontmatter}


\section{Introduction}

\begin{figure*}[t]
\centering
\includegraphics[width=\linewidth]{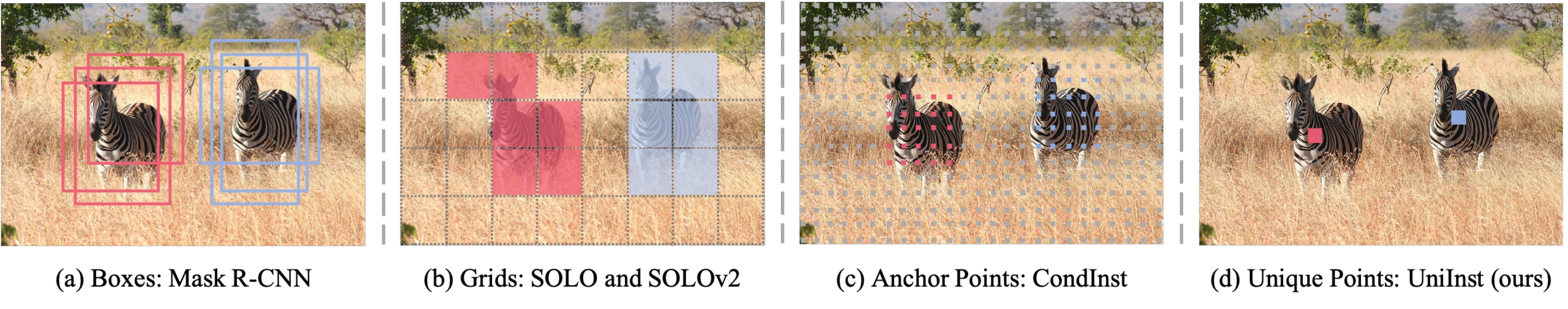}
\caption{Comparisons of different representation forms. Previous methods represent instances in different ways, such as \textbf{(a)} Bounding boxes of detected instances~\cite{he2017mask}, \textbf{(b)} Adjacent grids where instances locate~\cite{wang2020solo, wang2020solov2} or \textbf{(c)} Anchor points that hit the central region of instances~\cite{CondInst}. During inference, (a), (b), and (c) produce redundant representations and identify each instance by the NMS post-processing step. In our method, by contrast, \textbf{(d)} only one instance-aware \textbf{unique point} is yielded to one instance for end-to-end prediction without detection and post-processing.}
\label{fig:label_assign_comparison}
\end{figure*}

Instance segmentation is a fundamental yet challenging task in computer vision, which predicts a pixel-level mask and a semantic category for each instance in an image. Owing to the success of deep convolutional neural networks~\cite{simonyan2014very, szegedy2015going, he2016deep, chen2017deeplab, long2015fully}, instance segmentation has achieved impressive progress, with many well-performing approaches~\cite{he2017mask, wang2020solo, CondInst}. Among them, one challenging topic is how to represent instances. As illustrated in Figure~\ref{fig:label_assign_comparison} (a), (b), and (c), previous methods proposed to represent instances mainly via three forms. (a) Region-of-Interest-based (RoI-based) methods~\cite{he2017mask, huang2019mask, chen2019hybrid, kirillov2020pointrend, MaWDYLZ21} represent instances by boxes. They first employ an object detector to generate multiple bounding boxes for each instance and then crop features of boxes by RoI-Align~\cite{he2017mask} to predict instance masks. (b) SOLO~\cite{wang2020solo} and SOLOv2~\cite{wang2020solov2} represent instances through adjacent grids where the object locates. Then, they propose instance masks using center locations on S × S grids. (c) CondInst~\cite{CondInst} represents instances by anchor points landing in the center region of instances and predicts instance masks by dynamic weights subjected to these anchor points.

Although the above methods have achieved impressive performance, they still suffer from a challenging dilemma, wherein redundant representations, e.g., multiple boxes, grids, or anchor points, are assigned to one ground-truth instance (dubbed as the many-to-one assignment). As a result, all of them resort to non-maximum suppression (NMS) post-processing steps during inference, which is unsuitable for occluded as well as crowded scenarios and hinders the instance segmentation framework from end-to-end training. When eliminating NMS in mainstream instance segmentation methods~\cite{he2017mask, wang2020solo, wang2020solov2, CondInst}, results reported in Table~\ref{tab:assign} demonstrate that it is difficult to achieve satisfying performance dependent on the many-to-one assignment, e.g., 19.1 mask AP absolute drop on CondInst~\cite{CondInst}. 
One could intuitively address this issue by adding a mask head on top of end-to-end detectors, e.g., DETR~\cite{DETR} and DeFCN~\cite{DeFCN}.
Although these alternatives can offer a modest performance without the post-processing step (see Table~\ref{tab:assign}), they are not comparable to methods with the many-to-one assignment.
Additionally, they are extremely dependent on the object detector and are not complete end-to-end instance segmentation frameworks. To this end, one question may naturally arise: \textit{Could a fully convolutional network achieve a competitive and complete end-to-end instance segmentation framework?}

In this paper, we attempt to answer this question from the perspective of assignment between representations and ground-truth instances. We hereby propose a simple yet effective one-to-one assignment, namely \textbf{O}nly \textbf{Y}ield \textbf{O}ne \textbf{R}epresentation (\textbf{OYOR}), which dynamically assigns representations according to the matching quality between predictions and ground-truth instances. Specifically, as depicted in Figure~\ref{fig:label_assign_comparison} (d), only one unique representation with the highest matching quality is assigned as the positive sample to one ground-truth instance in our method, while others are suppressed effectively. The matching quality is evaluated by the classification confidence and the mask accuracy of each representation simultaneously, thus the proposed OYOR is instance-aware. Towards the unique representation of a ground-truth instance, we employ the unique instance-aware point and the corresponding dynamic weights to predict the mask of this instance. Therefore, our OYOR does not acquire box-based information and post-processing steps, which enables our framework box-free and NMS-free.

\begin{table}[t]
\centering
\caption{Comparisons with common instance segmentation methods on COCO \textit{val2017} split. Here, we adopt the ResNet-50-FPN backbone and $3\times$ schedule for all models except DETR. Mask R-CNN$^\ast$ is the improved Mask R-CNN by Detectron2~\cite{wu2019detectron2}. 'CondInst+DeFCN' refers to replacing the base detector of CondInst with the end-to-end DeFCN, and its pipeline shows in Figure~\ref{fig:pipeline} (c). AP and AR refer to mask mean average precision and mean average recall, respectively.}
\label{tab:assign}
\resizebox{\linewidth}{!}{
\begin{tabular}{c|c|c|ccc|ccc}
\hline
\multirow{2}{*}{Assignment types} &
  \multirow{2}{*}{Representation forms} &
  \multirow{2}{*}{Method} &
  \multicolumn{3}{c|}{AP} &
  \multicolumn{3}{c}{AR} \\ \cline{4-9} 
 &
   &
   &
  \multicolumn{1}{c|}{w/ NMS} &
  \multicolumn{1}{c|}{w/o NMS} &
  \multicolumn{1}{c|}{$\Delta$} &
  \multicolumn{1}{c|}{w/ NMS} &
  \multicolumn{1}{c|}{w/o NMS} &
  $\Delta$ \\ \hline
\multirow{4}{*}{Many-to-one} & Boxes     & Mask R-CNN$^*$~\cite{he2017mask}  & 37.2 & 10.3 & - 26.9 & 44.6 & 49.7  & + 5.1  \\
 & Grids     & SOLO~\cite{wang2020solo} & 35.8 & 17.3 & - 18.5 & 48.2 & 49.2 & + 1.0 \\
& Grids     & SOLOv2~\cite{wang2020solov2} & 37.6 & 17.9 & - 19.7 & 49.9 & 50.8 &  + 0.9 \\
& Anchor Points    & CondInst~\cite{CondInst} & 37.5 & 18.1 & - 19.4 & 48.7 & 52.2  & + 3.5 \\ \hline
\multirow{3}{*}{One-to-one}  & Queries & DETR~\cite{DETR} & --  & 31.9 & -- & -- & -- & -- \\
 & Boxes      & CondInst~\cite{CondInst}+DeFCN~\cite{DeFCN} & -- & 34.9 & -- & -- & -- & -- \\
 & Unique Points      & UniInst (\textbf{ours}) & \textbf{38.5} & \textbf{38.3} & -0.2 & \textbf{53.2} & \textbf{54.4} & +1.2 \\ \hline
\end{tabular}
}
\end{table}

Furthermore, existing methods~\cite{he2017mask, CondInst} directly adopt predicted classification scores as the unilateral ranking criterion to determine the final predictions in the inference stage. Consequently, the network may infer one sub-optimal instance mask with a high classification score but low mask quality as the prediction output. To mitigate this issue, we design \textbf{a prediction re-ranking strategy} to calibrate the ranking criterion to the product of classification score and mask quality. In this way, the predicted mask takes the instance quality into account, and the most discriminative and representative prediction is therefore inferred for each instance by our framework.

With proposed techniques, our complete end-to-end instance segmentation framework, termed \textbf{UniInst}, can directly perform a single mask prediction for each instance without procedures of detection and post-processing.
Experiments on COCO benchmark~\cite{COCO} show that our UniInst can achieve a competitive performance (40.2 mask AP with ResNet-101-FPN backbone) against mainstream NMS-based and query-based methods.
To further demonstrate its robustness and flexibility for occlusion scenes, we conduct additional experiments on the heavily-occluded OCHuman benchmark~\cite{zhang2019pose2seg}, where our method outperforms CondInst~\cite{CondInst} by a remarkable $+12.6$ mask AP.

Our main contributions can be summarized as follows:
\begin{itemize}
  \item We propose an effective one-to-one assignment scheme to prune redundant representations, equipping fully convolutional networks (FCNs) with the ability to learn an instance-aware representation for each instance uniquely in an end-to-end manner.
  
  \item A prediction re-ranking strategy is elegantly integrated into the end-to-end framework.  It calibrates the ranking criterion with mask quality and produces the most discriminative prediction that simultaneously considers the classification score and mask quality.
  
  \item Without relying on the detector and the post-processing step, our end-to-end UniInst achieves competitive performance on COCO \textit{test-dev2017}, e.g., 39.0 mask AP with ResNet-50-FPN backbone and 40.2 mask AP with ResNet-101-FPN backbone. Enabled by the instance-aware representation, our UniInst achieves superior performance (40.2 mask AP with ResNet-50-FPN backbone) on OCHuman dataset that contains more occlusion scenes.
\end{itemize}

\section{Related Work}
\subsection{Instance Segmentation}
Existing instance segmentation approaches can be roughly separated into two-stage and one-stage paradigms.
Two-stage methods~\cite{he2017mask, liu2018path, huang2019mask, chen2019hybrid, kirillov2020pointrend, cheng2020boundary} first employ object detectors to generate proposal boxes, then predict mask of each detected instance after RoI-Align~\cite{he2017mask}. Typically, Mask R-CNN~\cite{he2017mask} extends Faster R-CNN~\cite{ren2015faster} by adding an extra mask head. Based on~\cite{he2017mask}, Mask Scoring R-CNN~\cite{huang2019mask} explicitly learns the quality of predicted masks. HTC~\cite{chen2019hybrid} further improves Cascade R-CNN~\cite{cascade_mask} by interweaving box and mask branches in a multi-stage cascade manner. PointRend~\cite{kirillov2020pointrend} adaptively selects points to refine boundary details for image segmentation. 
One-stage methods~\cite{wang2020solo, wang2020solov2, CondInst, xie2020polarmask, zhang2020mask, chen2020blendmask, wu2020single, yang2022borderpointsmask, luo_crose} incorporate mask prediction into a single-shot FCN without RoI cropping. PolarMask~\cite{xie2020polarmask} represents the mask by its contour in polar coordinates and formulates the problem as distance regression. SOLO~\cite{wang2020solo} presents a box-free framework to map input images to full instance masks. SOLOv2~\cite{wang2020solov2} further decouples the mask prediction into dynamic weights and convolutional features learning. Similarly, CondInst~\cite{CondInst} takes advantage of dynamic weights to predict masks. Recently, Borderpointsmask~\cite{yang2022borderpointsmask} utilizes several boundary points to represent an instance's mask and boundary box without detection.
However, as mentioned above, these methods suffer from redundant representations inferred for each instance, requiring a hand-crafted post-processing. In this paper, we propose a fully end-to-end framework to directly perform a single prediction for each instance with the proposed one-to-one
assignment rule.

\subsection{Label Assignment}
Label assignment refers to defining the positive and negative samples. Generally, it can be summarized into two categories, including the many-to-one assignment and the one-to-one assignment.
The many-to-one assignment, widely used in mainstream object detectors~\cite{girshick2015fast,liu2016ssd,redmon2016you,lin2017focal,law2018cornernet,redmon2018yolov3,tian2019fcos,lin2017feature,ren2015faster}, refers to assigning many positive predictions for one ground-truth. Analogously, most instance segmentation methods follow the similar idea. Mask R-CNN~\cite{he2017mask} inherits multiple positive proposals generated from the region proposal network, then adopts box NMS post-processing for each instance. One-stage SOLO~\cite{wang2020solo}, SOLOv2~\cite{wang2020solov2} and CondInst~\cite{CondInst} adopt the center sampling strategy~\cite{tian2019fcos} for label assignment, where proposals in the center region region of instance are considered as positives. In inference, box or matrix NMS is used to suppress the redundant mask predictions. 
Recently, several multi-stage refinement detectors~\cite{DETR, zhu2020deformable, sun2020sparse,hu2018relation, chi2020relationnet++} present one-to-one assignment for object detection, where only one positive sample is assigned to one ground-truth. These methods perform single prediction for each instance, achieving comparable performance, but suffering from high computational overhead.

Different from these methods, we provide a new perspective to prune redundant representations in instance segmentation. Inspired by~\cite{DeFCN, sun2020onenet, hu2019statistical, hu2018adaptive}, we design a straightforward one-to-one assignment to dynamically assign one unique representation to one instance without post-processing. Besides, we design a novel prediction re-ranking strategy to help produce the most discriminative prediction.

\section{Method}
\begin{figure*}[t]
\centering
\includegraphics[width=\linewidth]{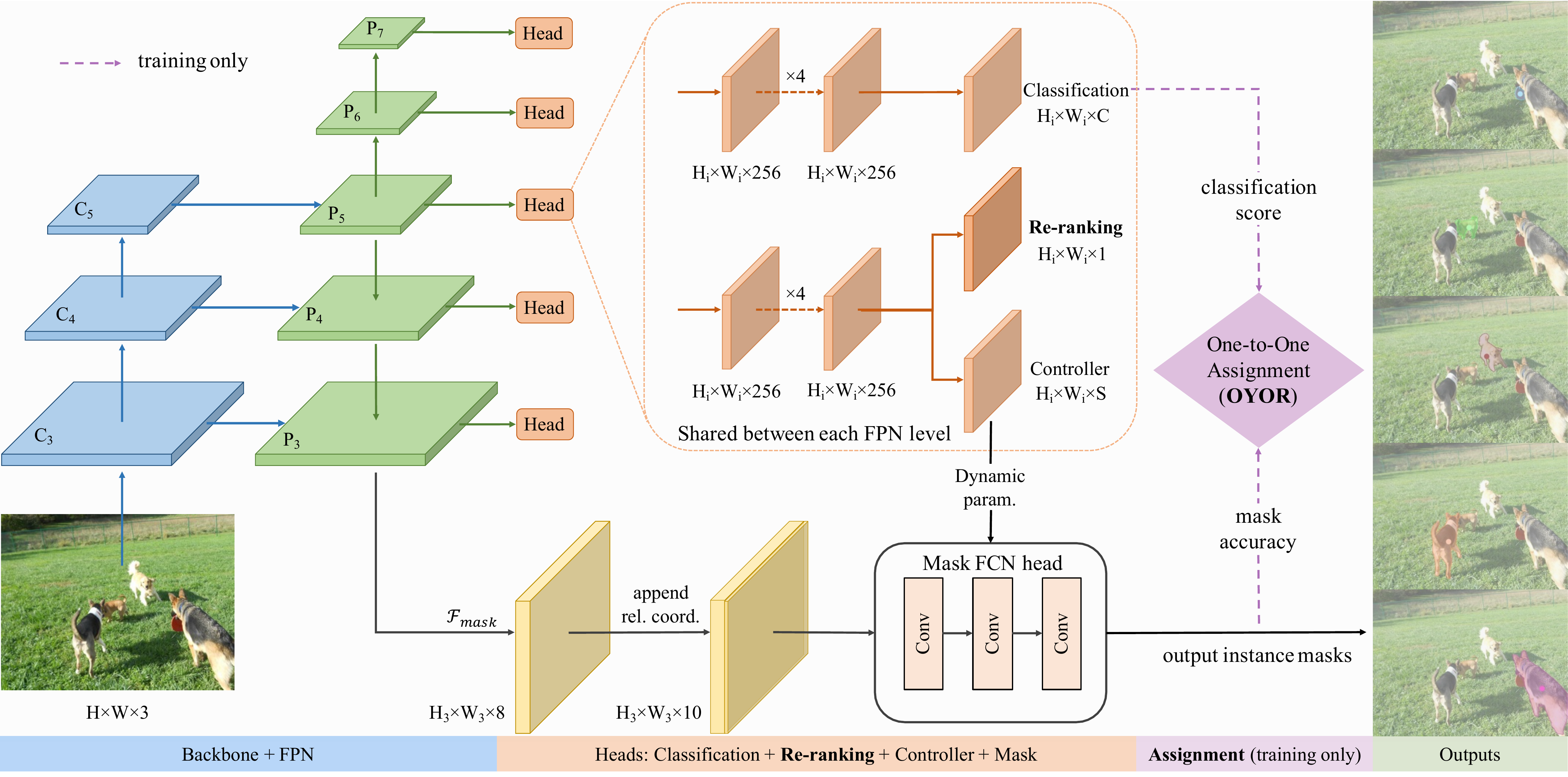}
\captionsetup{font=small}\caption{The diagram of the proposed UniInst. $C_3$ to $C_5$ are the feature maps from the backbone network. $P_3$ to $P_7$ are the feature maps from FPN~\cite{lin2017feature}, where the resolution of $P_i$ is $2^{i}$ times lower than that of the input. $\mathcal{F}_{mask}$ has the same resolution as $P_3$, which is appended relative coordinates and subsequently input into the mask head. The orange dashed heads are repeatedly applied on $P_3$ to $P_7$, where the Classification, Re-ranking (Sec.~\ref{subsec:re-ranking}), and Controller branches predict the class probability, mask IoU, and dynamic parameters, respectively. $S$ equals the total number of parameters in the mask FCN head. ‘OYOR’ indicates the proposed instance-aware one-to-one assignment scheme that enables entirely end-to-end instance segmentation (Sec.~\ref{subsec:oyor}). The purple dashed lines are used to highlight the additional operation in the training stage, which are abandoned in the inference stage.}
\label{fig:architecture}
\end{figure*}

In this section, we first empirically analyze the assignment scheme between representations and ground-truth instances. Then, we present an instance-aware one-to-one assignment scheme and a prediction re-ranking strategy, both methods enabling an end-to-end instance segmentation framework termed UniInst. Next, the overall pipeline of our UniInst, as illustrated in Figure~\ref{fig:architecture}, is introduced in detail. Finally, the main differences between our UniInst and other mainstream pipelines are discussed.

\subsection{Analysis on Representation Assignment}
As shown in Figure~\ref{fig:label_assign_comparison}, previous methods~\cite{he2017mask, wang2020solo, CondInst} adopt a many-to-one assignment scheme and subsequently require the NMS post-processing step to suppress redundant representations during inference. 
To demonstrate the effect of assignment scheme on instance segmentation, we conduct several ablation studies using mainstream methods~\cite{he2017mask, wang2020solo, CondInst} on COCO dataset~\cite{COCO}. As shown in Table~\ref{tab:assign}, when discarding NMS, due to the false-positive predictions generated from redundant representations, there is a dramatic performance drop for methods equipped with the many-to-one assignment scheme, e.g., 26.9 and 19.4 mask AP absolute drops on Mask R-CNN and CondInst, respectively. Therefore, it is challenging to achieve end-to-end learning through the many-to-one assignment solely. 
In order to avoid the post-processing step, an intuitive approach is to utilize the end-to-end detectors.
DETR~\cite{DETR}, relied on a one-to-one assignment scheme, can be equipped with a mask head on top of the decoder, which achieves 31.9 mask AP without the post-processing step.
We can also replace the detector of CondInst~\cite{CondInst} with the end-to-end DeFCN~\cite{DeFCN}, which performs 34.9 mask AP.
Without NMS, the above two substitute approaches surpass the many-to-one-based methods.
This phenomenon demonstrates the potentialities of the one-to-one assignment scheme on the NMS-free framework.
However, these two alternatives depend on detection results to predict instance masks, which is not a complete end-to-end instance segmentation framework. More importantly, the information from instance masks is exploited in the assignment deficiently. To mitigate this issue, we propose an effective one-to-one assignment scheme that dynamically assigns one unique representation to each instance according to the matching quality between predictions and ground-truth instances.

\subsection{UniInst}
\subsubsection{Instance-aware One-to-one Assignment: OYOR}
\label{subsec:oyor}
Let $\hat{y}$ be the set of all predictions and $y$ be the set of all ground truth instances. 
$N$ and $G$ correspond to the number of predictions and ground-truth instances, respectively, where $N$ is typically larger than $G$ in dense prediction frameworks. To achieve a one-to-one assignment between $N$ and $G$, we formulate a simple yet effective scheme, namely Only Yield One Representation (OYOR). 
It generates the optimal G-permutation of $N$ predictions from the perspective of bipartite matching.
As POTO of DeFCN~\cite{DeFCN}, OYOR uses the global matching quality instead of foreground loss~\cite{DETR} as the matching metric to alleviate optimization issues:
\begin{eqnarray}
\label{equ1}
\hat{\pi}&=&\mathop{\arg\max}_{\pi\in\Pi_G^N}\sum_{i=1}^{G}Q_{match}(\hat{y}_{\pi(i)},y_{i}),
\end{eqnarray}
where $\hat{\pi}$ denotes the optimal permutation with the highest quality in all permutations $\Pi_G^N$. 
$Q_{match}(\hat{y}_{\pi(i)},y_{i})$ is a pair-wise matching quality of the $\pi(i)$-th prediction $\hat{y}_{\pi(i)}$ with the $i$-th ground truth $y_{i}$.
In detail, the $i$-th ground truth can be seen as $y_{i}=(c_{i},m_{i})$, where $c_{i}$ and $m_{i}$ denote its target category and ground-truth mask, respectively. For the $\pi(i)$-th prediction $\hat{y}_{\pi(i)}=(\hat{p}_{\pi(i)},\hat{m}_{\pi(i)})$, $\hat{p}_{\pi(i)}$ and $\hat{m}_{\pi(i)}$ refer to its predicted classification scores and mask, respectively.
The matching quality is defined by the weighted geometric mean between the classification score and the mask accuracy: 
\begin{eqnarray}
\label{equ2}
Q_{match}(\hat{y}_{\pi(i)},y_{i})=
\underbrace{\mathbb I_{\lbrace\pi(i)\in\Psi_{i}\rbrace}}_{spatial\, prior}
\cdot\underbrace{{\left(\hat{p}_{\pi(i)}(c_{i})\right)}^{1-\alpha}}_{classification\,score}
\cdot \underbrace{{\left(Dice(\hat{m}_{\pi(i)},m_{i})\right)}^{\alpha}}_{mask\,accuracy},
\end{eqnarray}
where $\alpha\in\left[0,1\right]$ is a hyper-parameter that adjusts the relative importance between the classification score and the mask accuracy. $\alpha = 0.9$ is adopted by default, and more ablation studies are provided in Table~\ref{tab:alpha}.
$\Psi_{i}$ represents enabled prediction candidates by the widely used spatial prior~\cite{redmon2017yolo9000, lin2017focal, tian2019fcos, duan2019centernet}. The proposed OYOR adopts the center sampling strategy~\cite{tian2019fcos} to improve matching efficiency, in which only predictions hit in the central region of ground-truth masks are taken into account.
The central region is determined by the centroid of instance masks instead of the box center as POTO.
Note that the spatial prior used here only determines prediction candidates, not positive or negative samples. 
$\hat{p}_{\pi(i)}(c_{i})$ is the predicted classification score of the target class $c_i$.
Different from the detection-oriented POTO that utilizes box-based IoU as a quality metric, our OYOR is devised explicitly for the instance segmentation task.
Thus, to describe the mask from a finer level, we adopt the Dice similarity coefficient~\cite{milletari2016v} between predicted mask $\hat{m}_{\pi(i)}$ and the ground-truth mask $m_{i}$ as the mask accuracy.
The explicit form of the Dice similarity coefficient is given below:
\begin{equation}
\label{eq:dice}
Dice(\hat{m}_{\pi(i)},m_{i}) = \frac{2\cdot |\hat{m}_{\pi(i)}\cap m_{i}|}{|\hat{m}_{\pi(i)}|+|m_{i}| + \epsilon},
\end{equation}
in which $\epsilon$ is $1\times 10^{-5}$ by default.
$|\hat{m}_{\pi(i)}\cap m_{i}|$ refers to the intersection between $\hat{m}_{\pi(i)}$ and $m_{i}$, calculated as $\sum_{j=1}^{h\times w}\hat{m}_{\pi(i), j}\cdot m_{i,j}$ where $h$ and $w$ is the height and width of the mask. $|\hat{m}_{\pi(i)}|$ is calculated as $\sum_{j=1}^{h\times w}\hat{m}_{\pi(i), j}^{2}$. $|m_{i}|$ has the same routine.

The Hungarian algorithm~\cite{Hungarian, DeFCN, DETR} can rapidly calculate the best permutation $\hat\pi$ with the highest matching quality, namely the optimal one-to-one assignment, wherein the matching quality $Q_{match}(\cdot)$ leverages the spatial prior, classification score, and mask accuracy simultaneously.
As a result, the proposed OYOR assigns an unique instance-aware representation for each ground truth (see Figure~\ref{center_visual}). 
The computational complexity of the Hungarian algorithm is $\mathcal{O}(NG)$ for an input image with $N$ predictions and $G$ ground-truth instances.
Note that $N$ is acceptable because prediction candidates are limited by the center sampling strategy~\cite{tian2019fcos}.
Additionally, our OYOR only works during training and does not affect inference speed at all.

\subsubsection{Prediction Re-ranking Strategy}\label{rank}
\label{subsec:re-ranking}

To date, most instance segmentation methods directly adopt classification scores as a sole criterion for ranking predictions during inference. This means that output predictions are only determined by elements with top classification scores. However, classification response only serves for distinguishing the semantic category of proposals and is not aware of the actual mask quality. In this case, the network may assign one sub-optimal prediction with high classification scores but low mask quality as the unique representation, thus resulting in a drop in performance. 

To mitigate this issue, we propose a novel prediction re-ranking strategy to calibrate the ranking criterion with mask quality. Specifically, the mask Intersection-over-Union (IoU) is utilized to describe the quality of mask predictions. As shown in Figure~\ref{fig:architecture}, a compact re-ranking head is introduced to regress the predicted mask quality at all locations across different FPN~\cite{lin2017feature} levels.
During training, we take the mask IoU between the predicted instance mask $\hat{m}_{\pi(i)}$ and its matched ground-truth mask $m_{i}$,  denoted as $IoU(\hat{m}_{\pi(i)},m_{i})$, as the target of the re-ranking head. As formulated in Eq.~\ref{equ4}, the re-ranking loss $L_{rank}$ is only computed for predictions within the enabled candidate $\Psi_{i}$, and $L1$ loss is used to supervise the regressed IoUs.
\begin{equation}
\label{equ4}
L_{rank}= \mathbb I_{\lbrace\pi(i)\in\Psi_{i}\rbrace}\cdot\Vert \hat{IoU}_{\pi(i)}-IoU(\hat{m}_{\pi(i)},m_{i}) \Vert_{1},
\end{equation}
where $\hat{IoU}_{\pi(i)}$ refers to the predicted mask IoU for the $\pi(i)$-th predicted mask.
During inference, once we obtain the predicted IoUs, all instance predictions are properly re-ranked by multiplying the predicted mask IoUs and classification scores. Then, the re-ranked top predictions are output as final predictions. To this end, our re-ranking strategy ensures only the most discriminative and representative prediction will be inferred for each instance.

The proposed re-ranking head  comprises a single $3\times 3$ convolutional layer (stride=1). Its computational complexity is $\mathcal{O}(H_i W_i K^2 C_{in} C_{out})$, where $H_i$ and $W_i$ are the height and width of feature maps in the $i$-th head, respectively. The kernel size $K$ equals $3$. The number of input channels $C_{in}$ and output channels $C_{out}$ equals $256$ and $1$, respectively. In particular, the regression head ($C_{out}=4$) is removed in UniInst because the proposed OYOR takes full advantage of the instance information.
Thus, compared with CondInst~\cite{CondInst}, we can reduce the computational complexity during inference.

\subsubsection{Framework}
The proposed UniInst is developed based on the CondInst~\cite{CondInst} without center-ness and box regression branches, and further improved with the proposed instance-aware OYOR scheme and prediction re-ranking strategy.
The overall architecture is depicted in Figure~\ref{fig:architecture}. 

\vspace*{1\baselineskip}
\noindent \textbf{Backbone and Head.} Generally, we adopt the ResNet-50-FPN and ResNet-101-FPN ~\cite{he2016deep, zhao2017pyramid} as main backbones.
The Feature Pyramid Network (FPN) outputs multi-scale feature maps $\{P_3, ..., P_7\}$ where the resolution of $P_i$ is $2^i$ times lower than that of the input.
For the head network, we remove box regression and center-ness branches of CondInst~\cite{CondInst} because our UniInst attends to instance comprehensively. 
The head consists of parallel classification, re-ranking, and controller branches, which predict the classification probability, mask IoU, and dynamic parameters over all positions per multi-scale feature maps, respectively.
As CondInst~\cite{CondInst}, the features from FPN (denoted as $\mathcal{F}_{mask}$) are appended with relative coordinates and submitted to the Mask FCN head. For candidates determined by spatial prior, their corresponding dynamic weights are transferred into the Mask FCN head to predict masks.
While assigning samples to each ground truth, candidate predictions from center sampling are selected by the OYOR according to classification scores and mask accuracy.
Additionally, we adopt 3D Max Filtering (3DMF)~\cite{DeFCN} to improve the convolution discriminability in local region (related ablation study is given in Table~\ref{ablation-re-ranking}).

\vspace*{1\baselineskip}
\noindent \textbf{Overall Losses.} 
Although hybrid loss functions have been applied for different purposes in recent works~\cite{GOCERI2021104458, gocceri2021application}, we adopt the multi-task losses as:
\begin{equation}\label{equ5}
L = \lambda_{cls} \cdot L_{cls} + \lambda_{mask} \cdot L_{mask} + \lambda_{rank} \cdot L_{rank} + \lambda_{aux} \cdot L_{aux},
\end{equation}
where classification loss $L_{cls}$ and mask loss $L_{mask}$ are identical to~\cite{CondInst}. $L_{rank}$ is introduced in Eq.~\ref{equ4}. 
$L_{aux}$~\cite{DeFCN} indicates the many-to-one assignment auxiliary loss.
It is introduced to provide adequate supervision and enhance feature learning. In our UniInst, one ground truth only corresponds to one unique representation, which leads to less supervision for training. To cope with this, we adopt center sampling~\cite{tian2019fcos} with a slightly modified many-to-one assignment.
Concretely, we first compute the matching quality for each position and take positions with top-9 quality as candidates in each FPN level. Then, candidates whose quality over the average quality are assigned as positive, and Focal loss~\cite{lin2017focal} is employed for their supervision.
Note that the auxiliary loss adopted here are not necessary for our overall framework, only for enhancing feature learning. For simplicity, this auxiliary loss is adopted by default in the proposed UniInst, and related ablation studies are elaborated in Table~\ref{tab:ablation_aux}. $\lambda_{cls}=1$, $\lambda_{mask}=1$, $\lambda_{rank}=1$, and $\lambda_{aux}=1$ are balance weights for $L_{cls}$, $L_{mask}$, $L_{rank}$, and $L_{aux}$, respectively.

\vspace*{1\baselineskip}
\noindent \textbf{Pipeline.}
\begin{figure*}[t]
\includegraphics[width=\linewidth]{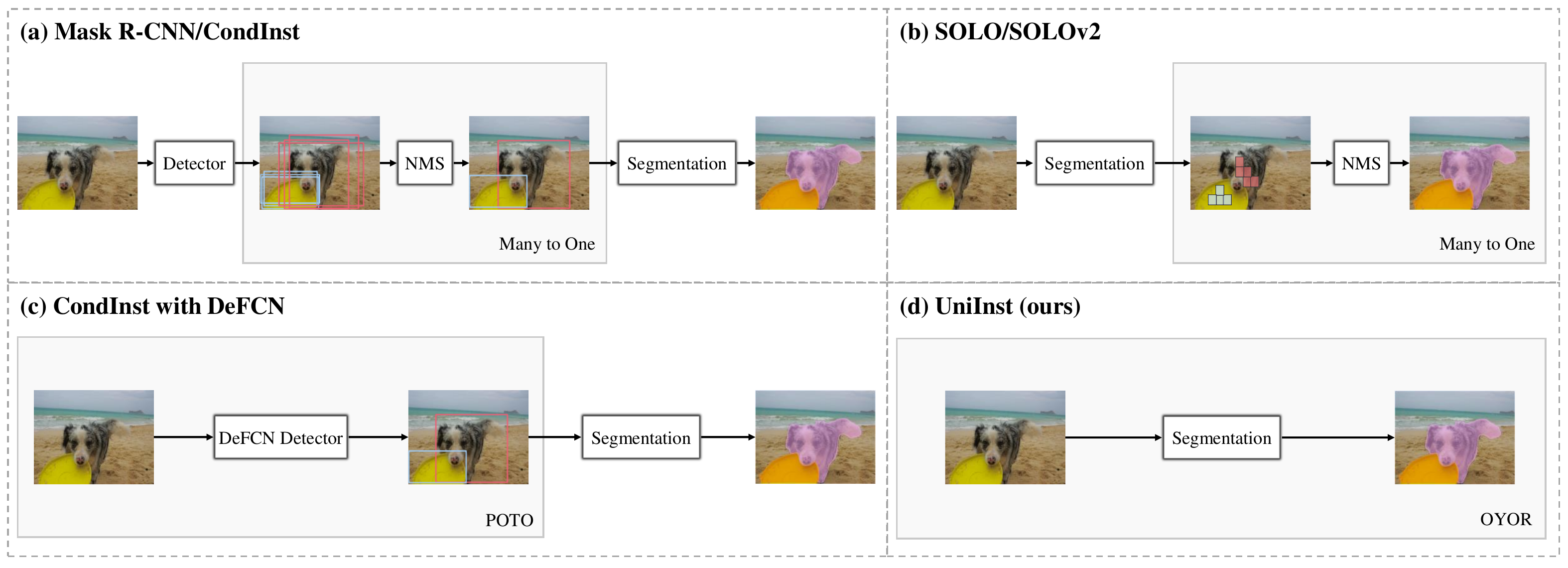}
\caption{Differences between our UniInst and mainstream instance segmentation methods in pipeline design.}
\label{fig:pipeline}
\end{figure*}
We design the pipeline of instance segmentation into a whole end-to-end style. As shown in Figure~\ref{fig:pipeline}, Mask R-CNN~\cite{he2017mask} and CondInst~\cite{CondInst} use a detector as a tool to identify instances. Thus, they are not only limited to the many-to-one assignments of the detector but also fail to use semantic level information to judge a positive sample.
Other works, e.g., SOLO~\cite{wang2020solo} and SOLOv2~\cite{wang2020solov2}, succeed in reaching comparable results without the detector, but they are still stuck into detection style assignment. Specifically, their pipelines do not utilize semantic level information to assign samples instead of grid center as multiple coarse assignments.
Figure~\ref{fig:pipeline} (c) shows an simple end-to-end instance segmentation approach by straightforwardly integrating the DeFCN~\cite{DeFCN} into the CondInst~\cite{CondInst}.
Wherein the assignment mechanism is POTO~\cite{DeFCN} that depends heavily on box-based information, including box center in center-sampling~\cite{tian2019fcos} and box-based IoU in judging instance quality (correspond to the first and third terms in Eq.~\ref{equ2}, respectively).
In contrast, the proposed OYOR utilize the centroid of instance masks to determine the center-sampling area and exploit the mask-based Dice IoU as a quality metric, which entirely depends on information from the fine-grained instance mask.
As a result, the UniInst achieves a streamlined box-free and NMS-free pipeline. The mask directly comes when an image passes our UniInst, as shown in Figure~\ref{fig:pipeline} (d). 

\section{Experiments}\label{Experiments}
In this section, we evaluate UniInst on COCO benchmark~\cite{COCO}, along with thorough comparisons and extensive ablation studies. 
To emphasize the robustness and flexibility of our method, we further conduct experiments on OCHuman~\cite{zhang2019pose2seg}, which contains more occluded and crowded scenes.

\subsection{Datasets}
\noindent \textbf{COCO:} COCO dataset~\cite{COCO} contains 118K images for training, 5K images for validation and 20K images for testing, involving 80 object categories with instance-level segmentation annotations. In this paper, we perform most of comparisons and ablations on COCO dataset. 
All models are trained on \textit{train2017} split, evaluated on \textit{val2017} split for ablation studies, and benchmarked on \textit{test-dev2017} split to compare with other methods.

\vspace*{1\baselineskip}
\noindent \textbf{OCHuman:} To further illustrate the effectiveness of UniInst in complex scenarios, i.e., occluded and crowded scenes, we perform test experiments on OCHuman~\cite{zhang2019pose2seg}, which is the most challenging dataset related to heavily-occluded humans. We selected 1761 accurately labeled images from OCHuman as a new benchmark, since the original dataset contains serious instances of missing annotations. Table~\ref{tab:tab_dataset} lists the “instance density” of different datasets, which shows that OCHuman contains more occluded and crowded scenes than COCO Person~\cite{COCO}, thus posing a big challenge for duplicate removal. 

\begin{table}[t]
\centering
\caption{Statistics of instance density on COCO Person~\cite{COCO} and OCHuman~\cite{zhang2019pose2seg}. The threshold for per image overlap statistics is ground-truth box IoU greater than $0.5$.}
\label{tab:tab_dataset}
\resizebox{0.6\textwidth}{!}{
\begin{tabular}{@{}c|c|c@{}}
\toprule
Datasets    & Images & Overlapped instances/image       \\ \midrule
COCO Person~\cite{COCO} & 2693   & 0.0204                           \\
OCHuman~\cite{zhang2019pose2seg}     & 1761   & \textbf{0.6417} \\ \bottomrule
\end{tabular}}
\end{table}

\subsection{Implementation Details}
\label{subsec:Implementation}
Our network is developed based on~\cite{CondInst}. Except for the new re-ranking head, all hyper-parameters are inherited from~\cite{tian2019fcos}. 
ResNet-50 and ResNet-101~\cite{he2016deep} with FPN~\cite{lin2017feature} are used as backbones. For a fair comparison, ResNet-50 and ResNet-101 are initialized by weights pre-trained on ImageNet and other new layers are initialized as~\cite{CondInst}.
Following~\cite{he2017mask, wang2020solo, CondInst}, input images are resized such that the shorter side is in [640, 800] and the longer side is less or equal to 1333 during training. During inference, the shorter side is set to 800. Following~\cite{he2017mask}, we train our models over 8 GPUs using stochastic gradient descent (SGD) with a mini-batch of 16 images (2 images per GPU) and an initial learning rate of 0.01.
To evaluate on COCO \textit{test-dev}, we train all models for 540K iterations (standard $6\times$ schedule), and the learning rate is reduced by a factor of 0.1 and 0.001 at iterations 480K and 520K, respectively. Unless otherwise stated, for ablation studies, we train all models for 270K iterations (standard $3\times$ schedule), and the learning rate is reduced at iterations 210K and 250K.

\subsection{Results on COCO}
\begin{figure*}[t]
\centering
\includegraphics[width=\linewidth]{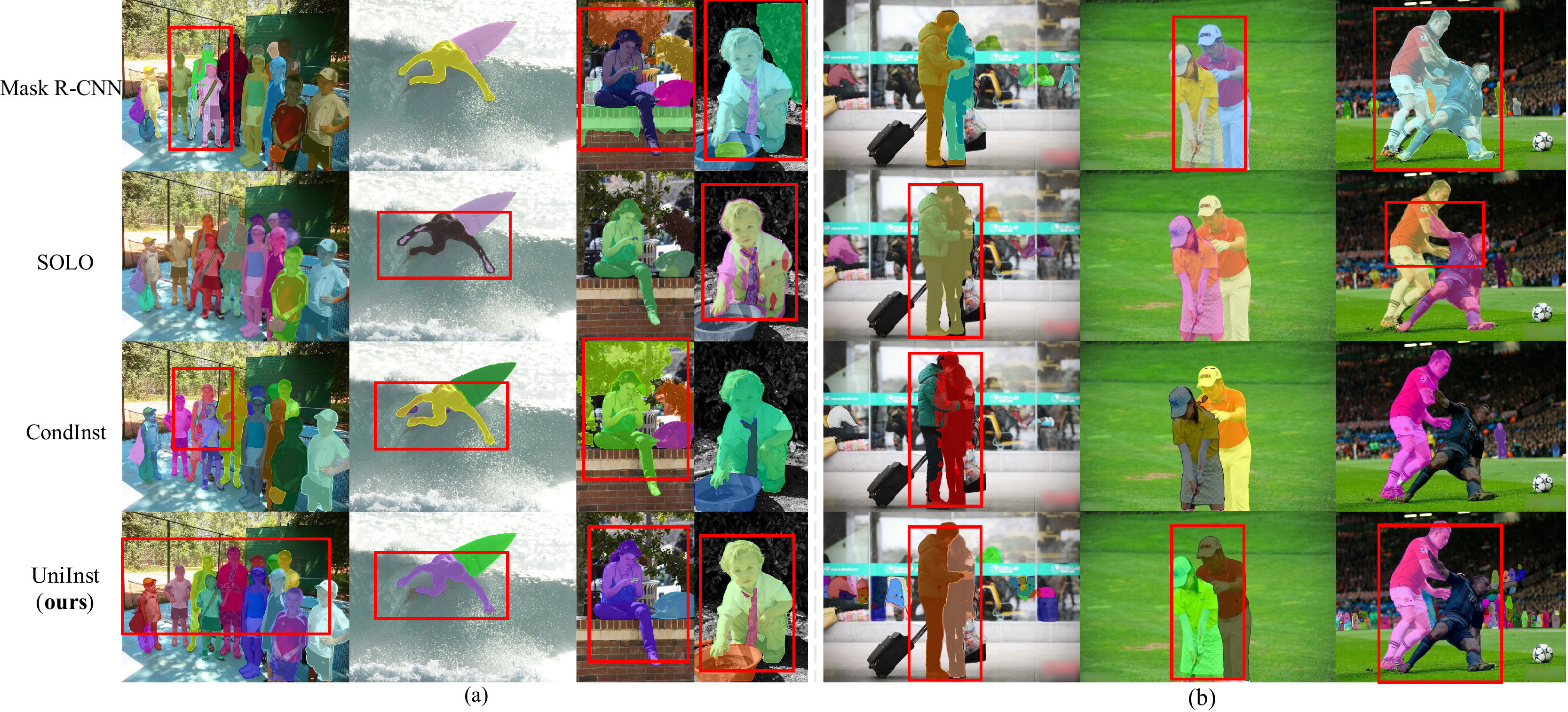}
\caption{Qualitative comparisons with other methods. We compare the proposed UniInst against Mask R-CNN~\cite{he2017mask}, SOLO~\cite{wang2020solo}, and CondInst~\cite{CondInst} on \textbf{(a)} COCO \textit{val2017} (left) and \textbf{(b)} OCHuman (right). Red boxes mark the areas that need to be focused on, where UniInst performs a high-quality instance segmentation, especially for scenes with occlusion.}
\label{fig:vis-comparison}
\end{figure*}

\begin{figure*}[h]
\centering
\includegraphics[scale=0.20]{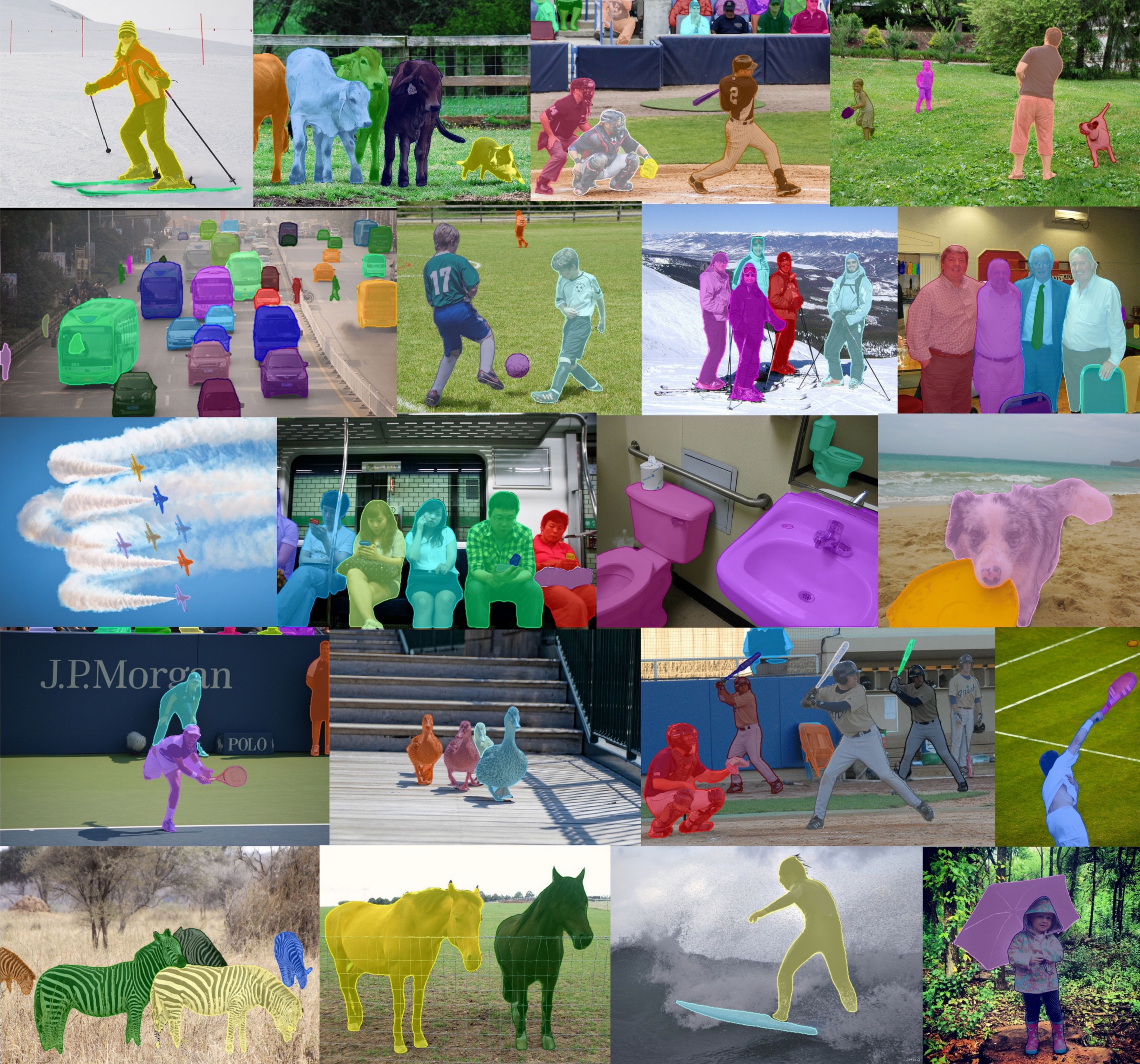}
\caption{More visualization of our UniInst with the ResNet-50-FPN backbone on COCO \textit{val2017} split.}
\label{fig:more-vis-coco}
\end{figure*}

We evaluate our UniInst with different backbones on COCO~\cite{COCO} dataset and compare it against mainstream methods. Table~\ref{tab:comparison_on_coco} shows that our UniInst obtains \textbf{39.0 mask AP} and \textbf{40.2 mask AP} with ResNet-50-FPN and ResNet-101-FPN backbones, respectively, achieving competitive performance against mainstream box-based and box-free methods. With ResNet-50-FPN, our UniInst outperforms the box-based Mask R-CNN and CondInst by $+1.5$ and $+1.2$ mask AP, respectively.
Compared with the box-free SOLO and SOLOv2, our UniInst obtains $+2.2$ and $+0.2$ mask AP gains, respectively. The UniInst also surpasses the query-based methods~\cite{K-Net} $+0.6$ mask AP. When based on the ResNet-101-FPN, similar performances are achieved by the proposed UniInst.
Figure~\ref{fig:vis-comparison} (a) shows some qualitative comparisons with the Mask R-CNN~\cite{he2017mask}, SOLO~\cite{wang2020solo}, and CondInst~\cite{CondInst}. It demonstrate that our UniInst performs a better segmentation for instances than other typical methods, especially for dense crowds. More qualitative results on COCO dataset are shown in Figure~\ref{fig:more-vis-coco}.

\begin{table*}[t]
\centering
\caption{Comparisons with state-of-the-art methods for the instance segmentation task on COCO \textit{test-dev2017}. 'Sched.' refers to the learning schedule. Mask R-CNN$^{*}$ means the model that improved by Detectron2~\cite{wu2019detectron2}. R-50 and R-101 denote ResNet-50 and ResNet-101, respectively. H-104 denotes Hourglass-104.}
\label{tab:comparison_on_coco}
\resizebox{1.0\textwidth}{!}{
\begin{tabular}{@{}l|c|c|ccc|ccc|c@{}}
\toprule
Method                                                            & Backbone       & Sched. & AP                   & AP$_{50} $ & AP$_{75}$ & AP$_{S}$ & AP$_{M}$ & AP$_{L}$ & Publication         \\ \midrule
\textbf{\textit{Box-based}}:    &                &        &                                &            &           &          &          &          &                     \\
Mask R-CNN~\cite{he2017mask}                      & R-50-FPN  & 1x     & 34.6                           & 56.5       & 36.6      & 15.4     & 36.3     & 49.7     & ICCV 2017           \\
Mask R-CNN$^*$~\cite{wu2019detectron2}                                                    & R-50-FPN  & 6x     & 37.5                           & 59.3       & 40.2      & 21.1     & 39.6     & 48.3     & ---           \\
TensorMask~\cite{chen2019tensormask}             & R-50-FPN  & 6x     & 35.4                           & 57.2       & 37.3      & 16.3     & 36.8     & 49.3     & ICCV 2019           \\
BlendMask~\cite{chen2020blendmask}               & R-50-FPN  & 3x     & 37.8                           & 58.8       & 40.3      & 18.8     & 40.9     & 53.6     & CVPR 2020           \\
Cascade Mask R-CNN~\cite{cascade_mask}          & R-50-FPN  & 3x     & 36.9                           & 58.6       & 39.7      & 19.6     & 39.3     & 48.8     & TIPAMI 2021           \\
HTC~\cite{chen2019hybrid}                         & R-50-FPN  & 3x     & 38.4                           & 60.0       & 41.5      & 20.4     & 40.7     & 51.2     & CVPR 2019           \\
CondInst~\cite{CondInst}                         & R-50-FPN  & 3x     & 37.8                           & 59.1       & 40.5      & 21.0     & 40.3     & 48.7     & TIPAMI 2022           \\
CondInst~\cite{CondInst}                        & R-50-FPN  & 6x     & 36.6                           & 57.4       & 39.0      & 18.8     & 39.3     & 47.9     & TIPAMI 2022           \\ \midrule
\textbf{\textit{Box-free}}:     &                &        &                                &            &           &          &          &          &                     \\
SOLO~\cite{wang2020solo}                          & R-50-FPN  & 6x     & 36.8                           & 58.6       & 39.0      & 15.9     & 39.5     & 52.1     & ECCV 2020           \\
SOLOv2~\cite{wang2020solov2}                    & R-50-FPN  & 6x     & 38.8                           & 59.9       & 41.7      & 16.5     & 41.7     & 52.6     & NIPS 2020           \\
K-Net (kernel-based)~\cite{K-Net}                         & R-50-FPN  & 3x     & 38.4 & 61.2 & 40.9       & 17.4      & 40.7     & 56.2 & NIPS 2021                  \\
UniInst (NMS-free, \textbf{ours})                                                 & R-50-FPN  & 6x     & \textbf{39.0} & 59.2       & 42.2      & 18.6     & 41.1     & 54.4     & Neurocomputing 2022                   \\ \midrule
\textbf{\textit{Box-based}}:    &                &        &                                &            &           &          &          &          &                     \\
Mask R-CNN~\cite{he2017mask}                     & R-101-FPN & 1x     & 35.7                           & 58.0       & 37.8      & 15.5     & 38.1     & 52.4     & ICCV 2017           \\
Mask R-CNN$^*$~\cite{wu2019detectron2}                                                    & R-101-FPN & 6x     & 38.8                           & 60.9       & 41.9      & 21.8     & 41.4     & 50.5     & ---           \\
TensorMask~\cite{chen2019tensormask}             & R-101-FPN & 6x     & 37.1                           & 59.3       & 39.4      & 17.4     & 39.1     & 51.6     & ICCV 2019           \\
BlendMask~\cite{chen2020blendmask}               & R-101-FPN & 6x     & 38.4                           & 60.7       & 41.3      & 18.2     & 41.5     & 53.5     & CVPR 2020           \\
Cascade Mask R-CNN~\cite{cascade_mask}          & R-101-FPN & 3x     & 38.4                           & 60.2       & 41.4      & 20.2     & 41.0     & 50.6     & TIPAMI 2021           \\
HTC~\cite{chen2019hybrid}                         & R-101-FPN & 3x     & 39.7                           & 61.8       & 43.1      & 21.0     & 42.2     & 53.5     & CVPR 2019           \\
CondInst~\cite{CondInst}                         & R-101-FPN & 3x     & 39.1                           & 60.9       & 42.0      & 21.5     & 41.7     & 50.9     & TIPAMI 2022           \\ \midrule
\textbf{\textit{Box-free}}:     &                &        &                                &            &           &          &          &          &                     \\
PolarMask~\cite{xie2020polarmask}                & R-101-FPN & 6x     & 30.4                           & 51.9       & 31.0      & 13.4     & 32.4     & 42.8     & CVPR 2020           \\
SOLO~\cite{wang2020solo}                         & R-101-FPN & 6x     & 37.8                           & 59.5       & 40.4      & 16.4     & 40.6     & 54.2     & ECCV 2020           \\
SOLOv2~\cite{wang2020solov2}                    & R-101-FPN & 6x     & 39.7                           & 60.7       & 42.9      & 17.3     & 42.9     & 57.4     & NIPS 2020           \\
CenterMask~\cite{wu2020single}                   & H-104  & 10.5x  & 34.5                           & 56.1       & 36.3      & 16.3     & 37.4     & 48.4     & ECCV 2020           \\
BorderPointsMask~\cite{yang2022borderpointsmask} & R-101-FPN & 1x     & 35.0                           & 56.5       & 37.1      & 17.1     & 37.4     & 48.6     & Neurocomputing 2022 \\
K-Net (kernel-based)~\cite{K-Net} & R-101-FPN & 3x     & 40.1                           & 62.8       & 43.1      & 18.7     & 42.7     & 58.8     & NIPS 2021 \\
UniInst (NMS-free, \textbf{ours})                                                 & R-101-FPN & 6x     & \textbf{40.2} & 61.0       & 43.6      & 19.4     & 42.8     & 55.9     & Neurocomputing 2022                   \\ \bottomrule
\end{tabular}}
\end{table*}

\subsection{Ablation Studies}

\subsubsection{Instance-aware One-to-one Assignment}
\label{ablation1}

\noindent \textbf{Representation Assignment.} To demonstrate the effect of representation assignment on discarding NMS, we conduct several ablation studies for mainstream methods~\cite{he2017mask, wang2020solo, CondInst}. As shown in Table~\ref{tab:assign} and Fig. \ref{fig:vis-comparison}(a), these methods are extremely sensitive to NMS. When discarding NMS, there is a dramatic drop in performance, $e.g.$, 19.4 mask AP absolute drop for CondInst~\cite{CondInst}. In contrast, our method only shows a very slight decrease (0.2 mask AP) and still outperforms NMS-based methods, which strongly demonstrates that our end-to-end framework has much fewer redundant representations than other methods.

\vspace*{1\baselineskip}
\noindent \textbf{Classification vs. Mask Accuracy.} The hyper-parameter $\alpha$ in Eq.~\ref{equ2} controls the ratio of mask accuracy and classification scores. As reported in Table~\ref{tab:alpha}, when $\alpha$ is 1, the gap with NMS is huge due to the misalignment between classification and mask prediction. When $\alpha$ is 0, the assignment scheme only relies on the predicted classification scores. In this case, the gap is considerably narrowed, but the overall performance is still not satisfactory. In contrast, with a proper fusion of classification and mask accuracy ($\alpha=0.9$), the performance is remarkably improved.

\vspace*{1\baselineskip}
\noindent \textbf{Pipeline Comparison.} To further demonstrate overall improvement against DeFCN~\cite{DeFCN}, we apply the concept of DeFCN directly to the instance segmentation domain. In Table~\ref{tab:assign}, “CondInst + DeFCN” means that we displace the detector of CondInst with DeFCN~\cite{DeFCN}, then following all implementations, including POTO assignment, 3D Max filter (3DMF), etc. However, this implementation still relies on a detector and exploits instance information inadequately. In comparison, our UniInst leverages instance properties and achieves higher mask AP than it (38.3 vs. 34.9 mask AP on COCO \textit{val2017} split).

\begin{table}[t]
\centering
\caption{Performance of UniInst with different configurations of $\alpha$ on COCO \textit{val2017} split.}
\label{tab:alpha}
\resizebox{0.28\textwidth}{!}{
\begin{tabular}{@{}c|ccc@{}}
\toprule
$\alpha$ & AP   & AP$_{50} $ & AP$_{75}$ \\ \midrule
0.0      & 33.5 & 53.5       & 35.4      \\
0.2      & 33.8 & 54.0       & 35.4      \\
0.4      & 34.4 & 54.6       & 36.6      \\
0.6      & 36.1 & 56.7       & 38.1      \\
0.8      & 37.1 & 57.8       & 39.3      \\
0.9      & \textbf{37.9} & \textbf{58.0}       & \textbf{40.9}      \\
1.0      & 11.8 & 17.2       & 12.8      \\ \bottomrule
\end{tabular}
}
\end{table}

\subsubsection{Prediction Re-Ranking Strategy}
We evaluate the effect of the proposed prediction re-ranking strategy and compare it with the 3DMF~\cite{DeFCN} and center-ness branch~\cite{tian2019fcos} in Table~\ref{ablation-re-ranking}.
The prediction re-ranking strategy and 3DMF bring $+1.2$ and $+0.8$ mask AP, respectively, while center-ness brings negative results ($-1.1$ mask AP). Note that improvements brought by the prediction re-ranking strategy are orthogonal with 3DMF. 
The proposed prediction re-ranking strategy still improves the performance by $+1.3$ mask AP when equipping with the 3DMF. Since it suppresses sub-optimal predictions with high classification scores but low mask accuracy, the final predictions aware of semantic categories and mask accuracy is achieved.

\begin{table}[t]
\centering
\caption{Ablation for 3DMF~\cite{DeFCN}, center-ness~\cite{tian2019fcos}, and our prediction re-ranking strategy on COCO \textit{val2017} split. All models are based on the ResNet-50-FPN backbone and trained $3\times$ learning schedule on COCO \textit{train2017} split.}
\label{ablation-re-ranking}
\resizebox{0.6\textwidth}{!}{
\begin{tabular}{ccc|ccc}
\toprule
3DMF & center-ness & re-ranking & AP & AP$_{50} $ & AP$_{75}$  \\
\midrule
 - &- &- & 35.6 & 56.0 & 38.2 \\
\checkmark &- & - & 36.4 & 57.2 & 39.0 \\
- &\checkmark & - & 33.5 & 51.7 & 36.0 \\
- &- & \checkmark & 36.8 & 56.2 & 39.6 \\
\checkmark& - & \checkmark  & \textbf{37.9} & \textbf{58.0} & \textbf{40.9} \\
\bottomrule
\end{tabular}}
\end{table}

\begin{figure*}[t]
\centering
\includegraphics[width=0.5\textwidth]{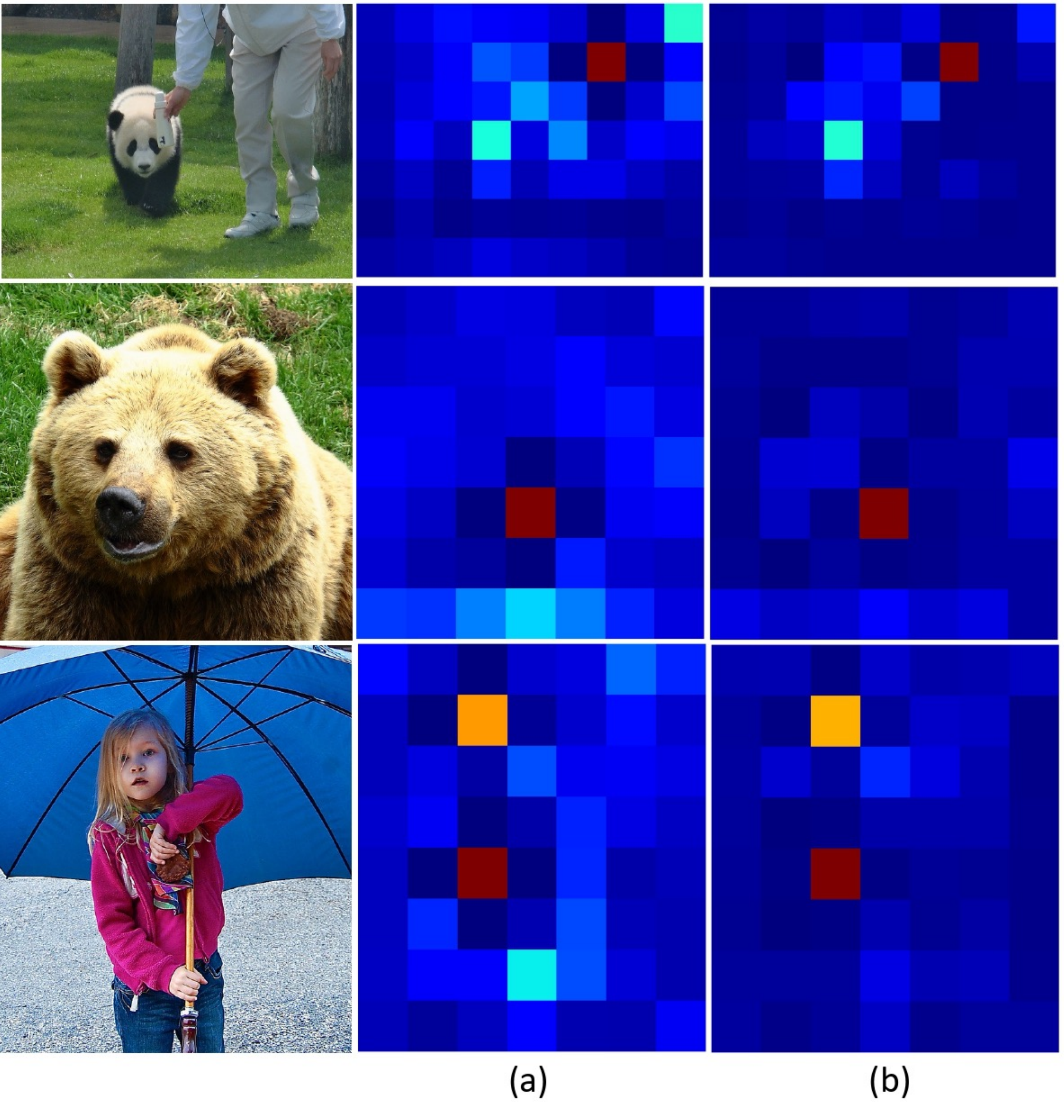}
\caption{Visualization of the predicted classification scores w/ and w/o the prediction re-ranking strategy on COCO \textit{val2017} split, shown in column (a) and (b), respectively.}\label{heatmap_visual}
\end{figure*}

We further visualize the classification scores during inference. As shown in Figure~\ref{heatmap_visual}, ranking with sole classification scores yields multiple predictions for single instance. These predictions are highly activated but have comparable scores with the most discriminative one. In this case, sub-optimal predictions with low mask quality are inferred. By contrast, when re-ranked by the predicted mask IoUs, these sub-optimal predictions are effectively suppressed, only predictions with high classification scores and mask quality are activated. As illustrated in Figure~\ref{heatmap_visual} (b), the learned feature is much sharper and discriminative.

\subsubsection{Auxiliary Loss}
We perform ablation studies to analyze the effect of the auxiliary loss adopted for optimization in Table~\ref{tab:ablation_aux}. 
Without this auxiliary loss, our approach works reasonably well, in which it still delivers a competitive performance (37.1 mask AP) against mainstream methods with NMS.
This performance indicates that the auxiliary loss is not necessary for the overall framework, but it can be beneficial for enhancing feature learning. We default this auxiliary loss to achieve better performance during training.
\begin{table}[t]
\centering
\caption{Ablation for auxiliary loss on COCO \textit{val2017} split. 'aux' refers to the auxiliary loss.}
\label{tab:ablation_aux}
\resizebox{0.42\textwidth}{!}{
\begin{tabular}{@{}l|ccc@{}}
\toprule
Method          & AP   & AP$_{50}$ & AP$_{75}$ \\ \midrule
UniInst w/o aux & 37.1 & 56.9   & 40.1   \\
UniInst w/ aux  & \textbf{37.9} & \textbf{58.0}   & \textbf{40.9}   \\ \bottomrule
\end{tabular}}
\end{table}

\subsubsection{Qualitative Visualization for the Unique Point}
\begin{figure}[t]
\centering
\includegraphics[width=\linewidth]{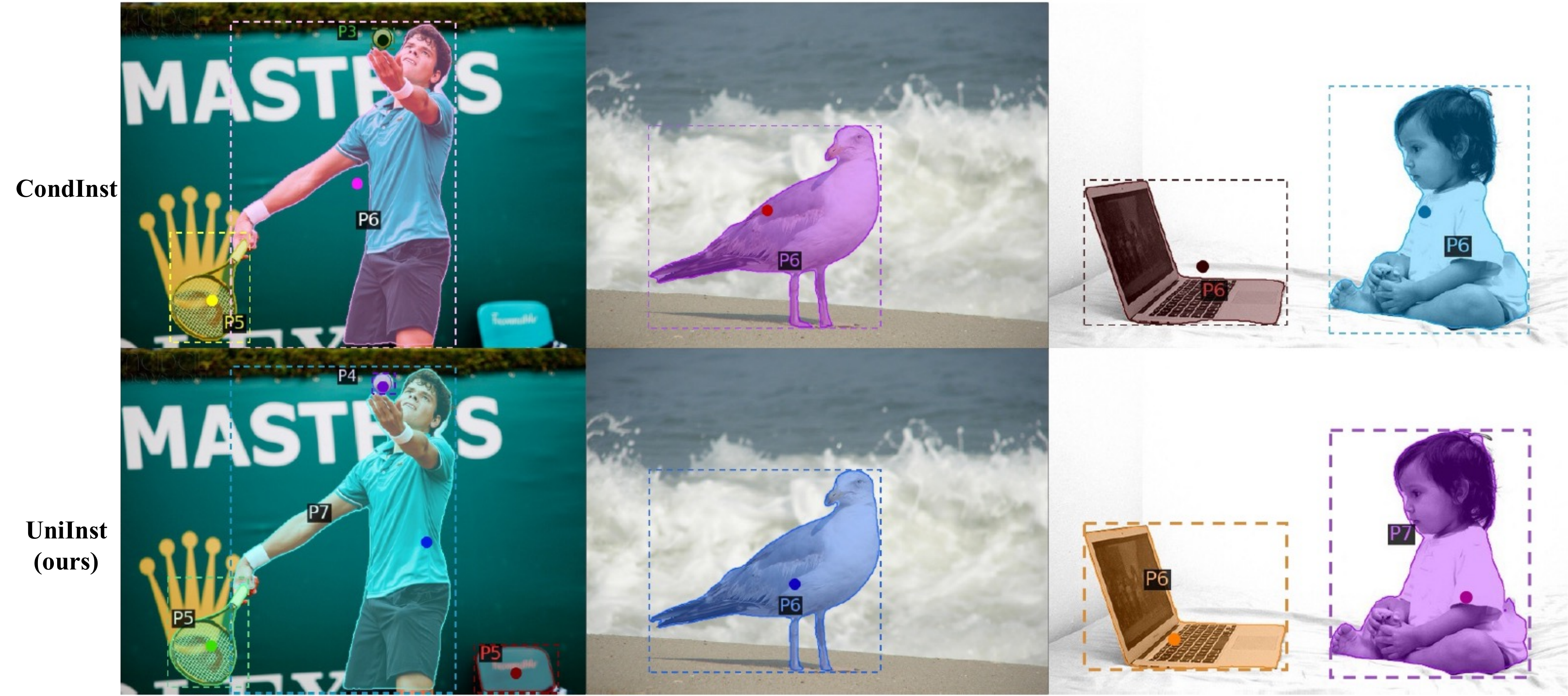}
\caption{Qualitative results of final predictions from CondInst~\cite{CondInst} and our UniInst with the ResNet-50-FPN backbone on COCO \textit{val2017} split. The dashed box is the outer rectangle of the instance mask, where the dark point is the location of the best representation point for the instance. CondInst tends to use the center point of the box to propose instance. By contrast, our UniInst utilize the instance-aware point to propose instance.}
\label{center_visual}
\end{figure}

As shown in Figure~\ref{center_visual}, we compare qualitative results of final predictions between CondInst~\cite{CondInst} and our UniInst. The main differences are: 1) CondInst~\cite{CondInst} follows the many-to-one assignment and NMS post-processing paradigm, while UniInst adopts the proposed instance-aware one-to-one assignment without any post-processing. 2) CondInst~\cite{CondInst} utilizes a center-ness branch to assist learning salient positions close to instance center, while UniInst uses the prediction re-ranking strategy to dynamically find the most discriminative point for each instance. Therefore, CondInst leverages NMS to find the best prediction that tends to lie in the grid point closest to ground-truth box center, but may fall out of instance, as the laptop in Figure~\ref{center_visual}. In contrast, the unique point from our UniInst (w/o NMS) exactly lies in the most discriminative region of instances, $e.g.$, inside of human body or laptop. It reveals that our UniInst is able to yield only one instance-aware unique representation for each instance without post-processing. 
Furthermore, we conduct the ablation study for the center-ness branch. As shown in Table~\ref{ablation-re-ranking}, a drop result demonstrates that the strong constraint of the center-ness is unsuitable for our instance-aware framework.

\begin{table}[t]
\centering
\caption{Comparisons with common FCN baseline methods for instance segmentation on OCHuman benchmark. Here, we adopt the $3\times$ learning schedule (36 epochs) and ResNet-50-FPN backbone for all models. AR denotes the mean average recall.}
\label{tab:OCHuman}
\resizebox{0.5\textwidth}{!}{
\begin{tabular}{@{}l|cccc@{}}
\toprule
Method                                                                    & AP                             & AP$_{50} $                     & AP$_{75}$                      & AR                             \\ \midrule
\multicolumn{1}{l|}{Mask R-CNN$^*$~\cite{wu2019detectron2}}                                       & 26.6                           & 65.9                           & 16.9                           & 40.5                           \\
\multicolumn{1}{l|}{SOLO~\cite{wang2020solo}}            & 37.5                           & \textbf{76.2} & 32.7                           & 53.1                           \\
\multicolumn{1}{l|}{CondInst~\cite{CondInst}} & 31.8                           & 69.9                           & 25.6                           & 47.6                           \\
\multicolumn{1}{l|}{UniInst (\textbf{ours})}                                              & \textbf{40.2} & 74.2                           & \textbf{38.9} & \textbf{55.8} \\ \bottomrule
\end{tabular}}
\end{table}

\subsection{Results on OCHuman}
\begin{figure*}[t]
\centering
\includegraphics[width=0.95\textwidth]{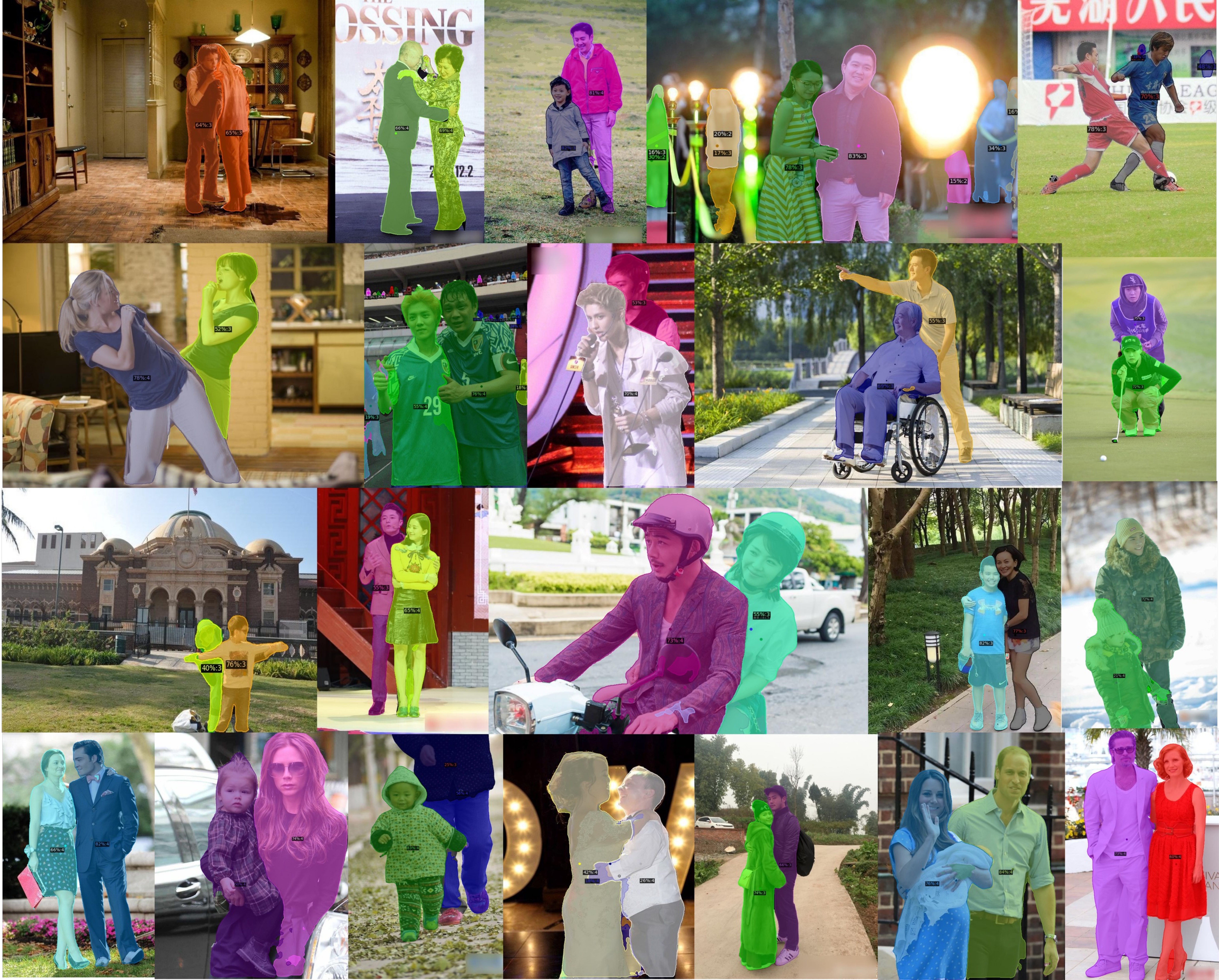}
\caption{More visualization of our UniInst with the ResNet-50-FPN backbone on OCHuman dataset.}
\label{more_oc}
\end{figure*}

To further illustrate the effect of instance-aware assignment (OYOR) and NMS-free from our UniInst, we perform experiments on complex OCHuman benchmark~\cite{zhang2019pose2seg}, which is the most challenging dataset related to heavily-occluded human. Following the same evaluation protocol in~\cite{zhang2019pose2seg}, our model is trained on general COCO \textit{train2017} split, and tested on OCHuman to evaluate its robustness instead of training on only occlusion cases. 
Table~\ref{tab:OCHuman} demonstrates that our UniInst shows great advantage in occluded scenes, outperforming other mainstream methods by remarkable mask APs, $e.g.$, 6.4 mask AP absolute gains over CondInst~\cite{CondInst}. 
Moreover, Figure~\ref{fig:vis-comparison}(b) illustrates the qualitative comparisons on OCHuman~\cite{zhang2019pose2seg}, where our UniInst obtains more accurate segmentation maps than other methods that fail to segment occluded instances. For example, the Mask R-CNN and SOLO are challenging to distinguish between two occluded persons. On the contrary, our UniInst can identify occluded persons obviously because of the instance-aware assignment and NMS-free framework. More qualitative results are shown in Figure~\ref{more_oc}.

\begin{table}[t]
\centering
\caption{Comparison of inference speed. All methods are based on ResNet-50-FPN backbone. The input size is same as the inference phase (provided in Sec.~\ref{subsec:Implementation}). The inference speed is measured on a single V100 GPU with 1 image per batch. AP refers the mask AP on COCO \textit{test-dev2017}. The training schedule is the same as Table~\ref{tab:comparison_on_coco}.}
\label{tab:speed}
\resizebox{0.7\textwidth}{!}{
\begin{tabular}{@{}l|c|cc@{}}
\toprule
Method              & AP & FPS ($img/s$) $\uparrow$ & Inf time ($ms/img$) $\downarrow$ \\ \midrule
Mask R-CNN$^*$~\cite{wu2019detectron2}& 37.5         & 17.5        & 57.1          \\
Cascade Mask R-CNN~\cite{cascade_mask}& 36.9  & 10.3        & 97.1          \\
SOLOv2~\cite{wang2020solov2} & 38.8            & 18.5        & 54.0          \\
CondInst~\cite{CondInst} & 37.8           & 20.4        & 49.0          \\
UniInst (\textbf{ours}) & \textbf{39.0}     & \textbf{21.1}      & \textbf{47.5}          \\
\bottomrule
\end{tabular}}
\end{table}

\subsection{Speed Analysis}
To demonstrate the practicality of our UniInst, we measure its efficiency and compare it with typical instance segmentation methods under the same conditions in Table~\ref{tab:speed}. The proposed UniInst takes 47.5 $ms$ to infer one image and process $\sim21$ images per second ($\sim21$ FPS). Its speed and accuracy gain the upper hand against Mask R-CNN$^*$~\cite{wu2019detectron2}, Cascade Mask R-CNN~\cite{cascade_mask} and SOLOv2~\cite{wang2020solov2}. 

\section{Discussion}
In this paper, we show that a fully convolutional network could achieve a competitive and complete end-to-end instance segmentation framework from the perspective of assignment. 
Nevertheless, there is still room for improvement in the future.
Firstly, results in Table~\ref{tab:comparison_on_coco} illustrate that the proposed UniInst struggles with segmenting small instances.
Future research should be devoted to the small instances segmentation from aspects of designing loss and assigning samples for small instances.
Secondly, each loss contributes equally to the overall loss in Eq.~\ref{equ5}. 
Adjusting the loss weights may affect the final results, but it is not the primary concern of this work. The AutoML~\cite{AutoML} can be utilized to search for the best loss weights in the future.
Additionally, time consumption from the convolutional and activation layers can be decreased by converting the model to half-precision or TensorRT format. However, the time spent on NMS post-processing, at least $\sim 2$ $ms$, cannot be reduced. Future work should pay more attention to the UniInst conversion and deployment.
As an extension, the proposed approach can be improved by the Transformer-based backbone~\cite{Swin, yang2022scalablevit} and compared with the performance of a capsule network-based approach~\cite{goceri2020capsnet, goceri2021analysis, goceri2021capsule}.


\section{Conclusion}
We have presented a novel end-to-end instance segmentation framework, UniInst, based on the fully convolutional network. With the proposed instance-aware one-to-one assignment scheme (OYOR) and the prediction re-ranking strategy, our box-free and NMS-free UniInst can yield the unique instance-aware point for each instance, thereby predicting a unique mask for each instance without the post-processing step. Extensive experiments on COCO benchmark demonstrate that our approach achieves effective and competitive performance against mainstream methods. Moreover, the proposed method is more robust to occlusion scenes, showing a great advantage on the heavily-occluded OCHuman benchmark. 
We wish the UniInst to pave the way for future research on the FCN-based end-to-end instance segmentation framework.

\section*{Acknowledgement}
This work was supported by the National Key R\&D Program of China (Grant No.2020AAA0108303), the National Natural Science Foundation of China (Grant No. 41876098)), and Shenzhen Science and Technology Project (Grant No. JCYJ20200109143041798).



\bibliographystyle{plainnat}
\bibliography{egbib.bib}

\begin{thebibliography}{58}
\providecommand{\natexlab}[1]{#1}
\providecommand{\url}[1]{\texttt{#1}}
\expandafter\ifx\csname urlstyle\endcsname\relax
  \providecommand{\doi}[1]{doi: #1}\else
  \providecommand{\doi}{doi: \begingroup \urlstyle{rm}\Url}\fi

\bibitem[Cai and Vasconcelos(2021)]{cascade_mask}
Zhaowei Cai and Nuno Vasconcelos.
\newblock Cascade {R-CNN:} high quality object detection and instance
  segmentation.
\newblock \emph{IEEE Transactions on Pattern Analysis and Machine
  Intelligence}, 43\penalty0 (5):\penalty0 1483--1498, 2021.

\bibitem[Carion et~al.(2020)Carion, Massa, Synnaeve, Usunier, Kirillov, and
  Zagoruyko]{DETR}
Nicolas Carion, Francisco Massa, Gabriel Synnaeve, Nicolas Usunier, Alexander
  Kirillov, and Sergey Zagoruyko.
\newblock End-to-end object detection with transformers.
\newblock In \emph{European Conference on Computer Vision, {ECCV}}, pages
  213--229, 2020.

\bibitem[Chen et~al.(2020)Chen, Sun, Tian, Shen, Huang, and
  Yan]{chen2020blendmask}
Hao Chen, Kunyang Sun, Zhi Tian, Chunhua Shen, Yongming Huang, and Youliang
  Yan.
\newblock Blendmask: Top-down meets bottom-up for instance segmentation.
\newblock In \emph{{IEEE/CVF} Conference on Computer Vision and Pattern
  Recognition, {CVRP}}, pages 8570--8578, 2020.

\bibitem[Chen et~al.(2019{\natexlab{a}})Chen, Pang, Wang, Xiong, Li, Sun, Feng,
  Liu, Shi, Ouyang, Loy, and Lin]{chen2019hybrid}
Kai Chen, Jiangmiao Pang, Jiaqi Wang, Yu~Xiong, Xiaoxiao Li, Shuyang Sun,
  Wansen Feng, Ziwei Liu, Jianping Shi, Wanli Ouyang, Chen~Change Loy, and
  Dahua Lin.
\newblock Hybrid task cascade for instance segmentation.
\newblock In \emph{{IEEE/CVF} Conference on Computer Vision and Pattern
  Recognition, {CVPR}}, pages 4974--4983, 2019{\natexlab{a}}.

\bibitem[Chen et~al.(2018)Chen, Papandreou, Kokkinos, Murphy, and
  Yuille]{chen2017deeplab}
Liang{-}Chieh Chen, George Papandreou, Iasonas Kokkinos, Kevin Murphy, and
  Alan~L. Yuille.
\newblock Deeplab: Semantic image segmentation with deep convolutional nets,
  atrous convolution, and fully connected crfs.
\newblock \emph{IEEE Transactions on Pattern Analysis and Machine
  Intelligence}, 40\penalty0 (4):\penalty0 834--848, 2018.

\bibitem[Chen et~al.(2019{\natexlab{b}})Chen, Girshick, He, and
  Doll{\'{a}}r]{chen2019tensormask}
Xinlei Chen, Ross~B. Girshick, Kaiming He, and Piotr Doll{\'{a}}r.
\newblock Tensormask: {A} foundation for dense object segmentation.
\newblock In \emph{{IEEE/CVF} International Conference on Computer Vision,
  {ICCV}}, pages 2061--2069, 2019{\natexlab{b}}.

\bibitem[Cheng et~al.(2020)Cheng, Wang, Huang, and Liu]{cheng2020boundary}
Tianheng Cheng, Xinggang Wang, Lichao Huang, and Wenyu Liu.
\newblock Boundary-preserving mask {R-CNN}.
\newblock In \emph{European Conference on Computer Vision, {ECCV}}, pages
  660--676, 2020.

\bibitem[Chi et~al.(2020)Chi, Wei, and Hu]{chi2020relationnet++}
Cheng Chi, Fangyun Wei, and Han Hu.
\newblock Relationnet++: Bridging visual representations for object detection
  via transformer decoder.
\newblock In \emph{Advances in Neural Information Processing Systems,
  {NeurIPS}}, 2020.

\bibitem[Duan et~al.(2019)Duan, Bai, Xie, Qi, Huang, and
  Tian]{duan2019centernet}
Kaiwen Duan, Song Bai, Lingxi Xie, Honggang Qi, Qingming Huang, and Qi~Tian.
\newblock Centernet: Keypoint triplets for object detection.
\newblock In \emph{{IEEE/CVF} International Conference on Computer Vision,
  {ICCV}}, pages 6568--6577, 2019.

\bibitem[Girshick(2015)]{girshick2015fast}
Ross~B. Girshick.
\newblock Fast {R-CNN}.
\newblock In \emph{{IEEE/CVF} International Conference on Computer Vision,
  {ICCV}}, pages 1440--1448, 2015.

\bibitem[Goceri(2020)]{goceri2020capsnet}
Evgin Goceri.
\newblock Capsnet topology to classify tumours from brain images and
  comparative evaluation.
\newblock \emph{IET Image Processing}, 14\penalty0 (5):\penalty0 882--889,
  2020.

\bibitem[Goceri(2021)]{GOCERI2021104458}
Evgin Goceri.
\newblock Diagnosis of skin diseases in the era of deep learning and mobile
  technology.
\newblock \emph{Computers in Biology and Medicine}, 134:\penalty0 104458, 2021.

\bibitem[G{\"O}{\c{C}}ER{\.I}(2021)]{gocceri2021application}
Evgin G{\"O}{\c{C}}ER{\.I}.
\newblock An application for automated diagnosis of facial dermatological
  diseases.
\newblock \emph{{\.I}zmir Katip {\c{C}}elebi {\"U}niversitesi Sa{\u{g}}l{\i}k
  Bilimleri Fak{\"u}ltesi Dergisi}, 6\penalty0 (3):\penalty0 91--99, 2021.

\bibitem[Goceri(2021{\natexlab{a}})]{goceri2021analysis}
Evgin Goceri.
\newblock Analysis of capsule networks for image classification.
\newblock In \emph{International Conference on Computer Graphics,
  Visualization, Computer Vision and Image Processing}, 2021{\natexlab{a}}.

\bibitem[Goceri(2021{\natexlab{b}})]{goceri2021capsule}
Evgin Goceri.
\newblock Capsule neural networks in classification of skin lesions.
\newblock In \emph{International Conference on Computer Graphics,
  Visualization, Computer Vision and Image Processing}, pages 29--36,
  2021{\natexlab{b}}.

\bibitem[He et~al.(2016)He, Zhang, Ren, and Sun]{he2016deep}
Kaiming He, Xiangyu Zhang, Shaoqing Ren, and Jian Sun.
\newblock Deep residual learning for image recognition.
\newblock In \emph{2016 {IEEE} Conference on Computer Vision and Pattern
  Recognition, {CVPR}}, pages 770--778, 2016.

\bibitem[He et~al.(2017)He, Gkioxari, Doll{\'{a}}r, and Girshick]{he2017mask}
Kaiming He, Georgia Gkioxari, Piotr Doll{\'{a}}r, and Ross~B. Girshick.
\newblock Mask {R-CNN}.
\newblock In \emph{{IEEE/CVF} International Conference on Computer Vision,
  {ICCV}}, pages 2980--2988, 2017.

\bibitem[He et~al.(2021)He, Zhao, and Chu]{AutoML}
Xin He, Kaiyong Zhao, and Xiaowen Chu.
\newblock Automl: {A} survey of the state-of-the-art.
\newblock \emph{Knowl. Based Syst.}, 212:\penalty0 106622, 2021.

\bibitem[Hu et~al.(2018{\natexlab{a}})Hu, Gu, Zhang, Dai, and
  Wei]{hu2018relation}
Han Hu, Jiayuan Gu, Zheng Zhang, Jifeng Dai, and Yichen Wei.
\newblock Relation networks for object detection.
\newblock In \emph{{IEEE/CVF} Conference on Computer Vision and Pattern
  Recognition, {CVPR}}, pages 3588--3597, 2018{\natexlab{a}}.

\bibitem[Hu et~al.(2018{\natexlab{b}})Hu, Monebhurrun, Himeno, Yokota, and
  Costen]{hu2018adaptive}
Runze Hu, Vikass Monebhurrun, Ryutaro Himeno, Hideo Yokota, and Fumie Costen.
\newblock An adaptive least angle regression method for uncertainty
  quantification in fdtd computation.
\newblock \emph{IEEE Transactions on Antennas and Propagation}, 66\penalty0
  (12):\penalty0 7188--7197, 2018{\natexlab{b}}.

\bibitem[Hu et~al.(2019)Hu, Monebhurrun, Himeno, Yokota, and
  Costen]{hu2019statistical}
Runze Hu, Vikass Monebhurrun, Ryutaro Himeno, Hideo Yokota, and Fumie Costen.
\newblock A statistical parsimony method for uncertainty quantification of fdtd
  computation based on the pca and ridge regression.
\newblock \emph{IEEE Transactions on Antennas and Propagation}, 67\penalty0
  (7):\penalty0 4726--4737, 2019.

\bibitem[Huang et~al.(2019)Huang, Huang, Gong, Huang, and Wang]{huang2019mask}
Zhaojin Huang, Lichao Huang, Yongchao Gong, Chang Huang, and Xinggang Wang.
\newblock Mask scoring {R-CNN}.
\newblock In \emph{{IEEE/CVF} Conference on Computer Vision and Pattern
  Recognition, {CVPR}}, pages 6409--6418, 2019.

\bibitem[Kirillov et~al.(2020)Kirillov, Wu, He, and
  Girshick]{kirillov2020pointrend}
Alexander Kirillov, Yuxin Wu, Kaiming He, and Ross~B. Girshick.
\newblock Pointrend: Image segmentation as rendering.
\newblock In \emph{{IEEE/CVF} Conference on Computer Vision and Pattern
  Recognition, {CVRP}}, pages 9796--9805, 2020.

\bibitem[Law and Deng(2018)]{law2018cornernet}
Hei Law and Jia Deng.
\newblock Cornernet: Detecting objects as paired keypoints.
\newblock In \emph{European Conference on Computer Vision, {ECCV}}, pages
  734--750, 2018.

\bibitem[Lin et~al.(2014)Lin, Maire, Belongie, Hays, Perona, Ramanan,
  Doll{\'{a}}r, and Zitnick]{COCO}
Tsung{-}Yi Lin, Michael Maire, Serge~J. Belongie, James Hays, Pietro Perona,
  Deva Ramanan, Piotr Doll{\'{a}}r, and C.~Lawrence Zitnick.
\newblock Microsoft {COCO:} common objects in context.
\newblock In \emph{European Conference on Computer Vision, {ECCV}}, pages
  740--755, 2014.

\bibitem[Lin et~al.(2017{\natexlab{a}})Lin, Doll{\'{a}}r, Girshick, He,
  Hariharan, and Belongie]{lin2017feature}
Tsung{-}Yi Lin, Piotr Doll{\'{a}}r, Ross~B. Girshick, Kaiming He, Bharath
  Hariharan, and Serge~J. Belongie.
\newblock Feature pyramid networks for object detection.
\newblock In \emph{{IEEE/CVF} Conference on Computer Vision and Pattern
  Recognition, {CVRP}}, pages 936--944, 2017{\natexlab{a}}.

\bibitem[Lin et~al.(2017{\natexlab{b}})Lin, Goyal, Girshick, He, and
  Doll{\'{a}}r]{lin2017focal}
Tsung{-}Yi Lin, Priya Goyal, Ross~B. Girshick, Kaiming He, and Piotr
  Doll{\'{a}}r.
\newblock Focal loss for dense object detection.
\newblock In \emph{{IEEE/CVF} International Conference on Computer Vision,
  {ICCV}}, pages 2999--3007, 2017{\natexlab{b}}.

\bibitem[Liu et~al.(2018)Liu, Qi, Qin, Shi, and Jia]{liu2018path}
Shu Liu, Lu~Qi, Haifang Qin, Jianping Shi, and Jiaya Jia.
\newblock Path aggregation network for instance segmentation.
\newblock In \emph{{IEEE/CVF} Conference on Computer Vision and Pattern
  Recognition, {CVPR}}, pages 8759--8768, 2018.

\bibitem[Liu et~al.(2016)Liu, Anguelov, Erhan, Szegedy, Reed, Fu, and
  Berg]{liu2016ssd}
Wei Liu, Dragomir Anguelov, Dumitru Erhan, Christian Szegedy, Scott~E. Reed,
  Cheng{-}Yang Fu, and Alexander~C. Berg.
\newblock {SSD:} single shot multibox detector.
\newblock In \emph{European Conference on Computer Vision, {ECCV}}, pages
  21--37, 2016.

\bibitem[Liu et~al.(2021)Liu, Lin, Cao, Hu, Wei, Zhang, Lin, and Guo]{Swin}
Ze~Liu, Yutong Lin, Yue Cao, Han Hu, Yixuan Wei, Zheng Zhang, Stephen Lin, and
  Baining Guo.
\newblock Swin transformer: Hierarchical vision transformer using shifted
  windows.
\newblock In \emph{{IEEE/CVF} International Conference on Computer Vision,
  {ICCV}}, pages 9992--10002, 2021.

\bibitem[Long et~al.(2015)Long, Shelhamer, and Darrell]{long2015fully}
Jonathan Long, Evan Shelhamer, and Trevor Darrell.
\newblock Fully convolutional networks for semantic segmentation.
\newblock In \emph{{IEEE/CVF} Conference on Computer Vision and Pattern
  Recognition, {CVPR}}, pages 3431--3440, 2015.

\bibitem[Luo et~al.(2021)Luo, Li, Gao, and Yan]{luo_crose}
Feng Luo, Xiu Li, Bin{-}Bin Gao, and Jiangpeng Yan.
\newblock A coarse-to-fine instance segmentation network with learning boundary
  representation.
\newblock In \emph{International Joint Conference on Neural Networks, {IJCNN}},
  pages 1--8, 2021.

\bibitem[Ma et~al.(2021)Ma, Wang, Dong, Yan, Li, and Zhang]{MaWDYLZ21}
Lufan Ma, Tiancai Wang, Bin Dong, Jiangpeng Yan, Xiu Li, and Xiangyu Zhang.
\newblock Implicit feature refinement for instance segmentation.
\newblock In \emph{2021 {ACM} International Conference on Multimedia, {MM}},
  pages 3088--3096, 2021.

\bibitem[Milletari et~al.(2016)Milletari, Navab, and Ahmadi]{milletari2016v}
Fausto Milletari, Nassir Navab, and Seyed{-}Ahmad Ahmadi.
\newblock V-net: Fully convolutional neural networks for volumetric medical
  image segmentation.
\newblock In \emph{Fourth International Conference on 3D Vision, {3DV}}, pages
  565--571, 2016.

\bibitem[Redmon and Farhadi(2017)]{redmon2017yolo9000}
Joseph Redmon and Ali Farhadi.
\newblock {YOLO9000:} better, faster, stronger.
\newblock In \emph{{IEEE/CVF} Conference on Computer Vision and Pattern
  Recognition, {CVRP}}, pages 6517--6525, 2017.

\bibitem[Redmon and Farhadi(2018)]{redmon2018yolov3}
Joseph Redmon and Ali Farhadi.
\newblock Yolov3: An incremental improvement.
\newblock \emph{arXiv preprint arXiv:1804.02767}, 2018.

\bibitem[Redmon et~al.(2016)Redmon, Divvala, Girshick, and
  Farhadi]{redmon2016you}
Joseph Redmon, Santosh~Kumar Divvala, Ross~B. Girshick, and Ali Farhadi.
\newblock You only look once: Unified, real-time object detection.
\newblock In \emph{{IEEE/CVF} Conference on Computer Vision and Pattern
  Recognition, {CVPR}}, pages 779--788, 2016.

\bibitem[Ren et~al.(2017)Ren, He, Girshick, and Sun]{ren2015faster}
Shaoqing Ren, Kaiming He, Ross~B. Girshick, and Jian Sun.
\newblock Faster {R-CNN:} towards real-time object detection with region
  proposal networks.
\newblock \emph{IEEE Transactions on Pattern Analysis and Machine
  Intelligence}, pages 1137--1149, 2017.

\bibitem[Simonyan and Zisserman(2015)]{simonyan2014very}
Karen Simonyan and Andrew Zisserman.
\newblock Very deep convolutional networks for large-scale image recognition.
\newblock In \emph{3rd International Conference on Learning Representations,
  {ICLR}}, 2015.

\bibitem[Stewart et~al.(2016)Stewart, Andriluka, and Ng]{Hungarian}
Russell Stewart, Mykhaylo Andriluka, and Andrew~Y. Ng.
\newblock End-to-end people detection in crowded scenes.
\newblock In \emph{{IEEE/CVF} Conference on Computer Vision and Pattern
  Recognition, {CVPR}}, pages 2325--2333, 2016.

\bibitem[Sun et~al.(2020)Sun, Jiang, Xie, Yuan, Wang, and Luo]{sun2020onenet}
Peize Sun, Yi~Jiang, Enze Xie, Zehuan Yuan, Changhu Wang, and Ping Luo.
\newblock Onenet: Towards end-to-end one-stage object detection.
\newblock \emph{arXiv preprint arXiv:2012.05780}, 2020.

\bibitem[Sun et~al.(2021)Sun, Zhang, Jiang, Kong, Xu, Zhan, Tomizuka, Li, Yuan,
  Wang, and Luo]{sun2020sparse}
Peize Sun, Rufeng Zhang, Yi~Jiang, Tao Kong, Chenfeng Xu, Wei Zhan, Masayoshi
  Tomizuka, Lei Li, Zehuan Yuan, Changhu Wang, and Ping Luo.
\newblock Sparse {R-CNN:} end-to-end object detection with learnable proposals.
\newblock In \emph{{IEEE/CVF} Conference on Computer Vision and Pattern
  Recognition,{CVPR}}, pages 14454--14463, 2021.

\bibitem[Szegedy et~al.(2015)Szegedy, Liu, Jia, Sermanet, Reed, Anguelov,
  Erhan, Vanhoucke, and Rabinovich]{szegedy2015going}
Christian Szegedy, Wei Liu, Yangqing Jia, Pierre Sermanet, Scott~E. Reed,
  Dragomir Anguelov, Dumitru Erhan, Vincent Vanhoucke, and Andrew Rabinovich.
\newblock Going deeper with convolutions.
\newblock In \emph{{IEEE/CVF} Conference on Computer Vision and Pattern
  Recognition, {CVPR}}, pages 1--9, 2015.

\bibitem[Tian et~al.(2022)Tian, Zhang, Chen, and Shen]{CondInst}
Z.~Tian, B.~Zhang, H.~Chen, and C.~Shen.
\newblock Instance and panoptic segmentation using conditional convolutions.
\newblock \emph{IEEE Transactions on Pattern Analysis and Machine
  Intelligence}, \penalty0 (01):\penalty0 1--1, 2022.
\newblock ISSN 1939-3539.

\bibitem[Tian et~al.(2019)Tian, Shen, Chen, and He]{tian2019fcos}
Zhi Tian, Chunhua Shen, Hao Chen, and Tong He.
\newblock {FCOS:} fully convolutional one-stage object detection.
\newblock In \emph{{IEEE/CVF} International Conference on Computer Vision,
  {ICCV}}, pages 9626--9635, 2019.

\bibitem[Wang et~al.(2021)Wang, Song, Li, Sun, Sun, and Zheng]{DeFCN}
Jianfeng Wang, Lin Song, Zeming Li, Hongbin Sun, Jian Sun, and Nanning Zheng.
\newblock End-to-end object detection with fully convolutional network.
\newblock In \emph{{IEEE/CVF} Conference on Computer Vision and Pattern
  Recognition, {CVPR}}, pages 15849--15858, 2021.

\bibitem[Wang et~al.(2020{\natexlab{a}})Wang, Kong, Shen, Jiang, and
  Li]{wang2020solo}
Xinlong Wang, Tao Kong, Chunhua Shen, Yuning Jiang, and Lei Li.
\newblock {SOLO:} segmenting objects by locations.
\newblock In \emph{European Conference on Computer Vision, {ECCV}}, pages
  649--665, 2020{\natexlab{a}}.

\bibitem[Wang et~al.(2020{\natexlab{b}})Wang, Zhang, Kong, Li, and
  Shen]{wang2020solov2}
Xinlong Wang, Rufeng Zhang, Tao Kong, Lei Li, and Chunhua Shen.
\newblock Solov2: Dynamic and fast instance segmentation.
\newblock In \emph{Advances in Neural Information Processing Systems,
  {NeurIPS}}, pages 17721--17732, 2020{\natexlab{b}}.

\bibitem[Wu et~al.(2020)Wu, Sahoo, Zhang, Zhu, and Hoi]{wu2020single}
Xiongwei Wu, Doyen Sahoo, Daoxin Zhang, Jianke Zhu, and Steven C.~H. Hoi.
\newblock Single-shot bidirectional pyramid networks for high-quality object
  detection.
\newblock \emph{Neurocomputing}, 401:\penalty0 1--9, 2020.

\bibitem[Wu et~al.(2019)Wu, Kirillov, Massa, Lo, and
  Girshick]{wu2019detectron2}
Yuxin Wu, Alexander Kirillov, Francisco Massa, Wan-Yen Lo, and Ross Girshick.
\newblock Detectron2.
\newblock \url{https://github.com/facebookresearch/detectron2}, 2019.

\bibitem[Xie et~al.(2020)Xie, Sun, Song, Wang, Liu, Liang, Shen, and
  Luo]{xie2020polarmask}
Enze Xie, Peize Sun, Xiaoge Song, Wenhai Wang, Xuebo Liu, Ding Liang, Chunhua
  Shen, and Ping Luo.
\newblock Polarmask: Single shot instance segmentation with polar
  representation.
\newblock In \emph{{IEEE/CVF} Conference on Computer Vision and Pattern
  Recognition, {CVRP}}, pages 12190--12199, 2020.

\bibitem[Yang et~al.(2022{\natexlab{a}})Yang, Zheng, Barzegar, Zhang, and
  Xu]{yang2022borderpointsmask}
Hanqing Yang, Liyang Zheng, Saba~Ghorbani Barzegar, Yu~Zhang, and Bin Xu.
\newblock Borderpointsmask: One-stage instance segmentation with boundary
  points representation.
\newblock \emph{Neurocomputing}, 467:\penalty0 348--359, 2022{\natexlab{a}}.

\bibitem[Yang et~al.(2022{\natexlab{b}})Yang, Ma, Wu, Tang, Xiao, Zheng, and
  Li]{yang2022scalablevit}
Rui Yang, Hailong Ma, Jie Wu, Yansong Tang, Xuefeng Xiao, Min Zheng, and Xiu
  Li.
\newblock Scalablevit: Rethinking the context-oriented generalization of vision
  transformer.
\newblock \emph{arXiv preprint arXiv:2203.10790}, 2022{\natexlab{b}}.

\bibitem[Zhang et~al.(2020)Zhang, Tian, Shen, You, and Yan]{zhang2020mask}
Rufeng Zhang, Zhi Tian, Chunhua Shen, Mingyu You, and Youliang Yan.
\newblock Mask encoding for single shot instance segmentation.
\newblock In \emph{{IEEE/CVF} Conference on Computer Vision and Pattern
  Recognition, {CVRP}}, pages 10223--10232, 2020.

\bibitem[Zhang et~al.(2019)Zhang, Li, Dong, Rosin, Cai, Han, Yang, Huang, and
  Hu]{zhang2019pose2seg}
Song{-}Hai Zhang, Ruilong Li, Xin Dong, Paul~L. Rosin, Zixi Cai, Xi~Han,
  Dingcheng Yang, Haozhi Huang, and Shi{-}Min Hu.
\newblock Pose2seg: Detection free human instance segmentation.
\newblock In \emph{{IEEE/CVF} Conference on Computer Vision and Pattern
  Recognition, {CVPR}}, pages 889--898, 2019.

\bibitem[Zhang et~al.(2021)Zhang, Pang, Chen, and Loy]{K-Net}
Wenwei Zhang, Jiangmiao Pang, Kai Chen, and Chen~Change Loy.
\newblock K-net: Towards unified image segmentation.
\newblock In \emph{Advances in Neural Information Processing Systems,
  {NeurIPS}}, pages 10326--10338, 2021.

\bibitem[Zhao et~al.(2017)Zhao, Shi, Qi, Wang, and Jia]{zhao2017pyramid}
Hengshuang Zhao, Jianping Shi, Xiaojuan Qi, Xiaogang Wang, and Jiaya Jia.
\newblock Pyramid scene parsing network.
\newblock In \emph{{IEEE/CVF} Conference on Computer Vision and Pattern
  Recognition, {CVRP}}, pages 6230--6239, 2017.

\bibitem[Zhu et~al.(2021)Zhu, Su, Lu, Li, Wang, and Dai]{zhu2020deformable}
Xizhou Zhu, Weijie Su, Lewei Lu, Bin Li, Xiaogang Wang, and Jifeng Dai.
\newblock Deformable {DETR:} deformable transformers for end-to-end object
  detection.
\newblock In \emph{9th International Conference on Learning Representations,
  {ICLR}}, 2021.

\end{thebibliography}


\end{document}